\documentclass[msom,nonblindrev]{informs3} 

\usepackage[top=1.0in,left=0.87in,right=0.87in,bottom=1.0in]{geometry}

\DoubleSpacedXI

\usepackage[draft]{hyperref}
\RequirePackage{fix-cm} 
\usepackage{xspace}
	\newcommand\ie{i.\,e.\xspace}
	\newcommand\eg{e.\,g.\xspace}

	\newcommand\US{U.\,S.\xspace}

	\usepackage{siunitx}
    \DeclareSIUnit\eur{\officialeuro}
    \DeclareSIUnit\M{M}
    \DeclareSIUnit\k{k}
	
	\usepackage[%
  sort&compress
]{cleveref}
  \crefname{chapter}{section}{sections}
	\Crefname{chapter}{Section}{Sections}

\usepackage{etoolbox}
\robustify\bfseries

\usepackage{url}
\urldef{\royalnursing}\url{https://www.goldstandardsframework.org.uk/cd-content/uploads/files/Primary%20Care/RCGP%20Matters%20of%20Life%20Death%20-%20Jul12.pdf}
\usepackage{isomath}
\usepackage{bm}

\usepackage{colortbl}
\usepackage{tabularx}
\usepackage{booktabs}
\usepackage{collcell}
\usepackage{subfig}

\usepackage{float}
\usepackage{rotating}

\usepackage{algorithm}
\usepackage{algpseudocode}

\usepackage{array}
\newcolumntype{L}[1]{>{\raggedright\let\newline\\\arraybackslash\hspace{0pt}}p{#1}}
\newcolumntype{C}[1]{>{\centering\let\newline\\\arraybackslash\hspace{0pt}}p{#1}}
\newcolumntype{R}[1]{>{\raggedleft\let\newline\\\arraybackslash\hspace{0pt}}p{#1}}

\usepackage{csquotes}
\usepackage{amsmath}
\usepackage{amssymb}
\usepackage{bbm}
\usepackage{pifont}
\usepackage[short]{optidef}

\usepackage[
    colorinlistoftodos,
    textsize=footnotesize,
        ]{todonotes}

\newcounter{daggerfootnote}

\makeatletter
    \renewcommand{\fps@figure}{H}         
    \renewcommand{\fps@table}{H}         
\makeatother 

\usepackage{float}
\usepackage{rotating}

\usepackage{enumitem}

\makeatletter
\newcolumntype{B}[3]{>{\boldmath\DC@{#1}{#2}{#3}}c<{\DC@end}}
\makeatother

\usepackage{natbib}
 \bibpunct[, ]{(}{)}{,}{a}{}{,}%
 \def\bibfont{\small}%
 \def\bibsep{\smallskipamount}%
\TheoremsNumberedThrough     

\EquationsNumberedThrough    

\MANUSCRIPTNO{MSOM-21-251} 

\interfootnotelinepenalty=10000

\begin{document}



\RUNTITLE{Data-Driven Allocation of Preventive Care}

\TITLE{Data-Driven Allocation of Preventive Care\\ With Application to Diabetes Mellitus Type II}

\ARTICLEAUTHORS{%
\AUTHOR{Mathias Kraus}
\AFF{FAU Erlangen-Nürnberg, Lange Gasse 20, 90403 Nürnberg, Germany, \EMAIL{mathias.kraus@fau.de}, \URL{}}
\AUTHOR{Stefan Feuerriegel}
\AFF{Munich Center for Machine Learning \& LMU Munich, Geschwister-Scholl-Platz 1, 80539 Munich, Germany, \EMAIL{feuerriegel@lmu.de}, \URL{}}
\AUTHOR{Maytal Saar-Tsechansky}
\AFF{Department of Information, Risk and Operations Management, McCombs School of Business, University of Texas at Austin, 2110 Speedway, B6000 Austin, TX 78712, USA, \EMAIL{maytal@mail.utexas.edu}, \URL{}}
} 

\ABSTRACT{%
\textbf{Problem Definition.} Increasing costs of healthcare highlight the importance of effective disease prevention. However, decision models for allocating preventive care are lacking.

\noindent\textbf{Methodology/Results.} In this paper, we develop a data-driven decision model for determining a cost-effective allocation of preventive treatments to patients at risk. Specifically, we combine counterfactual inference, machine learning, and optimization techniques to build a scalable decision model that can exploit high-dimensional medical data, such as the data found in modern electronic health records. Our decision model is evaluated based on electronic health records from \num{89191} prediabetic patients. We compare the allocation of preventive treatments (\emph{metformin}) prescribed by our data-driven decision model with that of current practice. We find that if our approach is applied to the \US population, it can yield annual savings of \$\num{1.1} billion. Finally, we analyze the cost-effectiveness under varying budget levels. 

\noindent\textbf{Managerial Implications.} Our work supports decision-making in health management, with the goal of achieving effective disease prevention at lower costs. Importantly, our decision model is generic and can thus be used for effective allocation of preventive care for other preventable diseases.
}


\HISTORY{}

\maketitle

\sloppy
\raggedbottom

\renewcommand\baselinestretch{0.77}
\selectfont

\vspace{-1cm}

\section{Introduction}
\label{sec:Introduction}


Millions of people die each year of diseases that could have been prevented. Common examples of preventable diseases include certain infectious diseases, such as human immunodeficiency virus (HIV); respiratory diseases, such as pneumonia; and cardiovascular diseases, such as diabetes mellitus. Over one quarter of all deaths in OECD countries are premature and could have been avoided through better preventive efforts \citep{oecd.2019}. Furthermore, the prevalence of preventable diseases is likely to rise as a result of current lifestyle choices \citep{WorldHealthOrganization.2018}. Preventable diseases reduce individuals' quality of life and also entail substantial costs. The cost of treating preventable diseases in the United States now totals \$730.4 billion annually, which corresponds to \SI{27.0}{\percent} of total healthcare spending \citep{Bolnick.2020}. 


Preventive care aims at managing risk factors before the onset of a disease \citep{WHO.2018}. A prominent example of preventive care for diabetes mellitus type~II is \emph{metformin}, which is a preventive drug aimed at controlling impaired glucose tolerance \citep{Knowler.2002}. However, many effective preventive treatments are often costly, and the effectiveness varies by risk factors \citep{Zhou.2020}. Consequently, healthcare managers face the challenge of cost-effective allocation of potentially limited or costly preventive treatments to patients. Indeed, with the growing importance of preventive care, effective allocation is essential in improving the quality of lives of billions of people worldwide. However, decision models for informing allocations in preventive care are lacking.  


Prior research has considered decision support for health management with the goal of achieving effective healthcare at lower costs. For instance, research has proposed optimizing periodic checkups \citep[\eg,][]{Ayvaci.2012, Ayvaci.2017, Kamalzadeh.2021, Liu.2018}, as well as the selection of treatment designs \citep[\eg,][]{Ibrahim.2016}. However, these works use generative models that are computationally prohibitive in the analysis of high-dimensional data to accurately predict future disease onsets. In contrast, discriminative machine learning models effectively scale to large datasets involving millions of observations and thousands of variables \citep[\eg,][]{Choi.2017}, which is common in modern healthcare settings. However, these models have not been adapted for allocation problems among patient populations in preventive care; this adaption is the objective of our work.  


Our work contributes to health management by proposing a data-driven decision model that provides decision support for allocation problems in preventive care. For this aim, we combine counterfactual inference, machine learning, and optimization techniques to exploit high-dimensional medical data from electronic health records~(EHRs). To our knowledge, this work contributes along three paths: (i)~we propose a data-driven decision model for allocating preventive treatments; (ii)~we demonstrate the value of scalable machine learning from rich, high-dimensional data in electronic health records for preventive care; and (iii)~we establish the \mbox{(cost-)}effectiveness of data-driven allocations. 


We evaluate our decision model for allocating preventive treatments for the case of diabetes mellitus type~II. Diabetes is a chronic disease that can significantly reduce patients' quality of life and that comes with substantial long-term costs to treat its many complications \citep{Lee.2018b}. For our evaluation, we use EHRs from \num{89191} patients with a prediabetic condition and analyze the allocation of preventive care in a dynamic setting spanning a period of 10 years. A simple, yet effective preventive treatment for diabetes type~II is \emph{metformin}. However, applying \emph{metformin} comes at a cost, thus raising the question of what a cost-effective allocation is in practice. We compare our decision model for preventive care allocation against a clinical baseline that stratifies preventive care according to the Framingham diabetes risk score. We find that our decision model prevents over \SI{25}{\percent} more cases of diabetes mellitus type II. Further, we find that, if our approach is applied to the \US population, it can yield annual savings of \$\num{1.1} billion because of a more effective allocation. 


Our decision model is directly relevant for healthcare organizations aimed at allocating preventive care under resource constraints. This is the case for vertically integrated healthcare organizations~(IHOs) worldwide, where hospitals and health insurances exchange patient data with the aim of achieving better coordination, efficiency, and quality of care for patients \citep{WHO.2016}. Examples of countries with IHOs are Israel, Singapore, many European countries, and partially the \US through health maintenance organizations \citep{WHO.2016}. 


Our work promises to have a direct effect on both practice and research. For healthcare practice, our research demonstrates that using a data-driven decision model based on scalable machine learning and EHRs can substantially improve the performance of preventive care programs and reduce the risk of disease onset. As such, our decision model supports health management in primary preventive care by offering a cost-effective path toward reduced health-related spending while improving patients' outcomes and quality of life. For research, our new decision model, incorporating counterfactual inference, offers a widely applicable solution for allocation problems in preventive care and for other allocation problems that require data-driven approaches in management science.


The remainder of our paper is structured as follows. \Cref{sec:background} provides an overview of preventive care, revealing a scarcity of decision models for this purpose. \Cref{sec:research_setting} details our research setting, which focuses on allocating preventive treatments to patients at risk of developing diabetes mellitus type II. In \Cref{sec:model_development}, we develop our data-driven decision model, combining counterfactual inference, machine learning, and optimization techniques to determine a cost-effective allocation of preventive care to patients at risk. We report our empirical results in \Cref{sec:empirical_results}. Finally, we discuss implications of our work for health management in \Cref{sec:discussion}.

\section{Background}
\label{sec:background}

Our work builds on literature in the fields of preventive care, machine learning, counterfactual inference, and resource allocation in health management. We review this work in the following sections. 

\subsection{Preventive Care}
\label{sec:preventive_care}

Preventive care refers to measures taken to reduce people's risk of developing a disease or suffering from a severe course of illness. Preventive care can be divided into three categories: primary prevention, secondary prevention, and tertiary prevention \citep{Goetzel.2009}. Primary prevention aims at averting the onset of a disease altogether. Examples include vaccines, smoking cessation, weight loss, and other measures that address risk factors through the promotion of good health \citep{WorldHealthOrganization.2018}. Secondary prevention focuses on the early stages of a disease after its onset. Here, the objective is to prevent a disease from developing into a critical or acute condition. Such approaches include screening (\eg, mammography) and other forms of early diagnosis. Tertiary prevention targets the progression associated with the disease. Here, the aim is to reduce the severity of symptoms. Examples include mental, physical, and social rehabilitation, such as in cardiac rehabilitation programs. In this work, we focus on primary prevention---specifically, on preventing patients from developing diabetes mellitus type II.

Preventive care can be a valuable economic investment; when it is provided early to patients, it can reduce the need for more expensive treatment after the onset of an illness \citep{WHO.2018}. Of note, the main aim in preventive care is to improve the health benefit for patients. Hence, optimization is measured against the overall health outcomes relative to costs, rather than against cost minimization alone, for ethical and legal reasons (see \citet{Gold.1996,NHS}, and \citet{WHO.2018}). However, not all patients provided with preventive care would necessarily have developed the disease in question. Given this premise, the cost-effectiveness of preventive care depends on two factors: (1) the expected number of people for whom diseases could be prevented relative to the population at risk and (2) the cost of providing preventive treatments. Preventive treatments, such as \emph{metformin}, can have low cost-effectiveness in cases where only a very small proportion of the population would have become ill without preventive care \citep{Cohen.2008}. In this case, targeting patients who are at high risk of developing a disease and for whom there is a large differential effectiveness of preventive efforts could potentially improve the overall cost-effectiveness.  

Previous research has assessed the cost-effectiveness of preventive care by comparing different preventive treatments or by comparing a preventive treatment against a placebo group \citep[\eg,][]{Breeze.2017,Cohen.2008, Zhou.2020}; but in these cases, the allocation rule was ex~ante given and thus not the focus of the study. For instance, in the context of prediabetic patients, the cost-effectiveness of preventive care (here, lifestyle coaching) was studied by allocating the preventive treatment to patients aged 40--65, with a body mass index~(BMI) higher than 35 and with blood glucose higher than \SI{6}{\percent} \citep{Breeze.2017}. However, the allocation rule used to determine when a prediabetic patient would receive the preventive care was devised by expert assessments and was not optimized through rigorous modeling. To fill this void, our work proposes a data-driven decision model.

\subsection{Machine Learning in Health Management}
\label{sec:data_driven_health_risk_scoring}

Machine learning is widely used to predict health outcomes and to adapt the decision-making in health management accordingly. Given the extensive body of research, we provide only a short summary in the following paragraphs and point toward \citet{Keskinocak.2020} for a detailed overview. In health operations, predictions make widespread use of EHRs as input. EHRs have emerged as part of the recent digitalization of healthcare operations \citep{KC.2020}; they encode a digital version of a patient's health trajectory, including a variety of risk variables, such as socio-demographics, body measurements (\eg, blood pressure), lab measurements, disease codes, and drug prescriptions. As a result, EHRs allow researchers to describe between-patient heterogeneity in health outcomes and, thereby, also enable a personalized view of a patient's individualized risk profile. Based on EHRs, health management can then predict different health outcomes. Examples of outcomes include the 30-day readmission risk as a performance indicator in hospital operations \citep[\eg,][]{Bardhan.2015}, mortality \citep[\eg,][]{Bjarnadottir.2018}, the onset of a disease \citep{Choi.2017}, or complications \citep[\eg,][]{Schallmoser.2023}. In this work, we use machine learning for predicting a patient's risk of developing diabetes mellitus type II. 

To predict the onset of diseases, research has used a variety of machine learning methods \citep[\eg,][]{Allam.2021, Bertsimas.2016, Zueger.2022}. Examples include lasso \citep{tibshirani1996regression}, ridge regression \citep{hoerl1970ridge}, random forests \citep{breiman2001random}, neural networks \citep{lecun2015deep}, and gradient boosted decision trees \citep{friedman2001greedy}. Throughout this work, we use gradient boosted decision trees as our main model because this machine learning model has shown promising results in other management applications \citep[\eg,][]{Glaeser.2019, Senoner.2021}. We later assess other machine learning methods as part of a sensitivity analysis.

\subsection{Counterfactual Inference in Health Management}
\label{sec:counterfactual_inference_in_health_management}

In health management, counterfactual inference is used to estimate the effects of different treatment options on health outcomes. Such estimates, in turn, inform the treatment choice, so that the treatment with the best expected effect on a patient's trajectory is chosen \citep{Bica.2021}. The gold standard for estimating treatment effects is a randomized controlled trial, although it also has several shortcomings \citep{Imbens.2015}. First, treatment effects may vary across patient populations, meaning that randomized controlled trials must be tailored to the patient population of interest. Second, randomized controlled trials are costly. Because of the immense upfront costs, performing such trials was not a viable course of action for our partnering IHO. Third, randomized controlled trials typically build on only two (or very few) treatment arms. As such, the estimated treatment effects typically are population-wide averages, with little personalization. Recent works \citep[\eg,][]{Bica.2021} offer a remedy for these shortcomings by using machine learning for estimating treatment effects from observational data, such as EHRs.

Previous works in counterfactual inference have pursued varying objectives \citep{Guo.2020}. On the one hand are studies that make inferences at the group level and thus estimate average treatment effects \citep[e.g.,][]{frauen2023estimating,vanderLaan.2006,Shi.2019}. On the other hand are studies that estimate heterogeneous treatment effects based on a patient's risk score \citep{Bica.2021}. Here, the assumption is that there is no global treatment effect of similar size for all patients but a differential treatment effectiveness---that is a heterogeneous treatment effect that varies across patients. As such, these studies may consider heterogeneous treatment effects related to, for instance, age, sex, or other risk scores. In the case of diabetes, preventive treatments such as \emph{metformin} are known to have a differential effectiveness \citep{Knowler.2002}. Thus, in our work, we use counterfactual inference to estimate heterogeneous treatment effects of \emph{metformin} from EHRs.

We build on a recent method called causal forest \citep{Wager.2018}. The causal forest is a non-parametric method that extends the widely used random forest algorithm, by \citet{breiman2001random}, for estimating heterogeneous treatment effects. Causal forests inherit several favorable properties from random forests in that they are expressive, require little tuning, and have a low risk of overfitting, thus offering a robust performance in practice. Further, under common mathematical assumptions, causal forests have been shown to be pointwise consistent for the true treatment effect and thus lead to provably valid inferences \citep{Wager.2018}.

\subsection{Resource Allocation in Health Management}
\label{sec:resource_allocation_in_health_management}

Research on health management examines allocation problems related to various resources. For an overview, we refer to \citet{Dai.2020} and \citet{Keskinocak.2020}. Examples include beds \citep[\eg,][]{Helm.2011}, drugs \citep{Khademi.2015}, aid \citep{Jakubik.2022}, and medical devices \citep{Deo.2015}. Other studies have used data-driven modeling for scheduling the admission of patients, so that available resources are used effectively but without exceeding surge capacities \citep[\eg,][]{May.2011}. Importantly, such problems generally optimize allocations according to patient needs (rather than focusing only on monetary aspects), so that providing \mbox{(cost-)}effective care is intended to improve patient health and not only to minimize costs. Analogously, we also focus on \mbox{(cost-)}effective care.  


Allocation decisions have been extensively modeled in the context of secondary prevention, with the goal of optimizing screening or check-up policies. Examples include biopsy referral \citep{Ayvaci.2017}, colonoscopy screening \citep{Erenay.2014}, mammography \citep{Cevik.2018}, and post-discharge monitoring \citep{Helm.2016, Liu.2018}. In this work, the main question is the optimal timing for patient examinations \citep{Kamalzadeh.2021}. Therefore, Markov decision processes or variants thereof typically are used in which (i)~decisions for an individual patient are made and (ii)~where states represent the patient's current health status. However, our problem setup for primary preventive care is different in both aspects: (i)~decisions are made for a cross-sectional sample (i.e., determining which patients out of a cohort should be selected for treatment), and (ii)~the future health trajectory should be used for decision-making, which requires machine learning models for prediction. 

Ethical and legal frameworks around the world mandate whom to allocate treatments, including preventive treatments, given scarce resources \citep[\eg,][]{NHS}. Accordingly, the core principle for health management is to focus on the best health benefit for patients, relative to the estimated costs, as opposed to prioritizing cost minimization alone. Hence, decisions in medicine are mainly benchmarked according to their cost-effectiveness \citep{Gold.1996}: Decision models should optimize against the largest improvement in health outcomes, given budget constraints (typically set by policymakers or healthcare organizations).

\section{Research Setting}
\label{sec:research_setting}

\subsection{The Case of Diabetes Prevention}
\label{sec:the_case_of_diabetes_prevention}

Our research aims at allocation of preventive care to patients at risk of developing diabetes mellitus type II. Diabetes mellitus is a chronic condition and, although widespread, is often preventable. In the United States, 88 million adults, or \SI{34.5}{\percent} of the adult population, are classified as prediabetic and thus at risk of developing diabetes if it is not prevented \citep{U.S.CentersforDiseaseControlandPrevention.2020}. According to the \citet{WHO.2018b}, diabetes mellitus not only is among the top 10 leading causes of death, but also can seriously imped the quality of life for patients in the long run. However, the risk of diabetes mellitus can be effectively mitigated with preventive treatments, including the drug \emph{metformin} \citep{Knowler.2002}. Once \emph{metformin} has been prescribed and then taken for diabetes prevention, patients generally must remain on the medication for the rest of their lives.


In this research, we partnered with a national health insurer from Israel serving more than 1.1 million patients. In Israel, health insurers are vertically integrated healthcare organizations, which enables our partnering health insurer to have direct access to a wealth of medical data from EHRs, as well as to work collaboratively with healthcare providers to deliver care to patients. We later discuss the applicability of our model to vertically integrated healthcare organizations from other countries (see \Cref{sec:discussion}) and, for that reason, refer to our partner company simply as IHO.

Our IHO has set a goal of improving primary preventive care for patients at risk of developing diabetes mellitus type II to mitigate individuals' risk of onset. For this, we develop and evaluate a data-driven allocation of preventive care using \emph{metformin}. Formally, the IHO is confronted by a decision-making problem: which of its patients $i = 1, \ldots, N$ should be enrolled in preventive treatment ($t_i = 1$) and which ones should not ($t_i = 0$)? If patient $i$ is enrolled in preventive care, her risk of developing diabetes is reduced by a treatment effect, which we later estimate from historical EHRs. However, the IHO has limited the overall budget available for preventive care to $k \leq N$ treatments. 

For primary prevention, the objective is to find a cost-effective allocation---that is, to maximize the number of prevented disease onsets, given a particular budget \citep{WHO.2018}. As with other research in health management \citep[\eg,][]{Ayvaci.2012,Helm.2011,Helm.2016}, the question is how to offer effective care at lower costs and thus to allocate resources for improving patient outcomes (as opposed to a goal of pure cost minimization, without considering health outcomes). Therefore, the objective in our work is to identify an allocation of preventive treatments to patients that maximizes the expected number of prevented onsets. To support the IHO in this task, we develop a data-driven decision model. 

In addition to looking at the number of prevented disease onsets, we also report costs, thus allowing us to study cost-effectiveness. Note that we intentionally refrain from having only a cost minimization focus because of ethical and legal principles \citep{NHS} and, as is common in medicine \citep{Gold.1996}, we focus on cost-effectiveness. Depending on the success of preventive treatments, the IHO incurs different costs. A preventive treatment incurs an annual cost of $C_\text{prevent}$ per patient. By enrolling all $N$ patients in the preventive program, the IHO would experience a total cost of $N \times C_\text{prevent}$, which generally would exceed available financial resources. If the disease is not prevented, the IHO faces annual costs of treating a patient with diabetes, $C_\text{diab}$, which depends both on the age of the patient and on any concomitant diseases caused by the developed diabetes. The cost for treatment usually is substantially larger than the cost for prevention---that is, $C_\text{diab} \gg C_\text{prevent}$. However, not all prediabetic patients actually develop diabetes; hence, the cost $C_\text{diab}$ occurs only for a subset of the overall study population.     

\subsection{Electronic Health Records}


Our work builds on an extensive longitudinal dataset. Specifically, our partnering IHO provided us with the EHRs of all patients classified as prediabetic and thus at risk of developing diabetes mellitus type II. Our sample comprises \num{89191} patients and exhibits considerable between-patient heterogeneity in risk profiles. For purposes of this research, access to the EHRs was granted for the time period of 2003 through 2012. 


For each patient, the EHRs comprise variables from the following categories: (i)~socio-demographics; (ii)~body measurements; (iii)~lab tests; (iv)~disease codes; and (v)~drug prescriptions. Disease codes provide data on other co-occurring conditions and are encoded based on the International Classification of Diseases (ICD) system of diagnostic codes. \Cref{tbl:variables} shows examples for each category (i)--(v). These variables present potential risk factors that describe heterogeneity in disease onsets among patients, and we thus leverage them later for machine learning. 

\begin{table}
    \TABLE
	{Variables in our dataset. \label{tbl:variables}}
	{\scriptsize
	\begin{tabular}{l r l}
		\toprule
		\textbf{Category} & \textbf{Number of variables} & \textbf{Examples} \\
		\midrule
		Socio-demographics & 2 & Age, sex \\
		Body measurements & 5 & Height, weight, body mass index,\\
		& & systolic blood pressure, diastolic blood pressure \\
		Lab tests & 100 & HbA1c, HDL-cholesterol, fasting glucose, \ldots \\
		Disease codes & 100 & Hypertensive diseases, disorders of metabolism, \ldots \\
		Drug prescriptions & 100 & Metformin, tritace, cardiloc, \ldots \\
		\midrule
		Overall ($=$input for our decision model) & 307 \\
		\bottomrule
	\end{tabular}}
	{}
\end{table} 


Patients were classified into (a)~{{prediabetic}} or (b)~{{diabetic}} as follows: (a)~Prediabetic is the precursor stage before the onset of diabetes mellitus type II. Following established criteria, patients are considered prediabetic when the fasting glucose level is between 6.1 mmol/L and 6.9 mmol/L \citep{WHO.2006}. The IHO used this inclusion criterion when compiling our patient data. (b)~Diabetic corresponds to a fasting glucose level above 6.9 mmol/L \citep{WHO.2006}. Later, during machine learning, our target variable denotes whether a patient has transitioned from prediabetic to diabetic ($=1$) or not ($=0$).

\subsection{Summary Statistics}


Our dataset comprises \num{89191} prediabetic patients. Of these patients, \num{77036} remain prediabetic throughout the study period, whereas \num{12155} develop diabetes mellitus. Hence, only \SI{13.62}{\percent} of patients transition from prediabetes to diabetes during the study period. \Cref{tbl:descriptives} lists summary statistics for key variables for the first record of each patient in our dataset. In our study population, the average age is 47.07, and the average body mass index (BMI) is 29.05. These observations are in line with earlier findings from medical research, according to which (pre-)diabetes is especially prevalent in elderly patients and patients with high BMI \citep{U.S.CentersforDiseaseControlandPrevention.2020}. 

Further differences arise when comparing patients who developed diabetes vs. patients who stayed prediabetic. We find that the average age is lower among patients who did not experience onset (46.84 years), compared to patients who developed diabetes (50.62 years). Similarly, the average BMI is larger for patients who developed diabetes (30.52 kg/m$^2$) compared to patients who stayed prediabetic (28.95 kg/m$^2$). Both observations are in line with medical research, which has identified age and BMI as important risk factors for the onset of diabetes \citep{U.S.CentersforDiseaseControlandPrevention.2020}. 

\begin{table}
	\TABLE
	{Summary statistics of example variables (for the first record of each patient in our dataset).\label{tbl:descriptives}}
	{\scriptsize
	\begin{tabular}{l SS SS SS}
		\toprule
		\textbf{Variable} & \multicolumn{2}{c}{\textbf{Overall}} & \multicolumn{2}{c}{\textbf{Without onset}} & \multicolumn{2}{c}{\textbf{With onset}} \\
		\cmidrule(lr){2-3}\cmidrule(lr){4-5}\cmidrule(lr){6-7}
		& \textbf{Mean} & \textbf{SD} & \textbf{Mean} & \textbf{SD} & \textbf{Mean} & \textbf{SD} \\ 
		\midrule
		Age (in years) & 47.07 & 17.93 & 46.84 & 18.32 & 50.62 & 9.28  \\
        Sex (0=male; 1=female) & 0.51 & 0.50 & 0.50 & 0.50 & 0.54 & 0.50  \\
        Systolic blood pressure (in mm\,Hg) & 122.84 & 17.11 & 122.54 & 16.51 & 127.43 & 24.01  \\
        Diastolic blood pressure (in mm\,Hg) & 77.26 & 12.78 & 77.16 & 12.93 & 78.80 & 10.12  \\
        Body mass index (in kg/m$^2$) & 29.05 & 5.64 & 28.95 & 5.62 & 30.52 & 5.73  \\
        HbA1c (in \%) & 5.56 & 0.36 & 5.51 & 0.34 & 5.87 & 0.34  \\
		\bottomrule
		\multicolumn{6}{l}{SD = standard deviation} \\
	\end{tabular}}
	{}
\end{table}  

\subsection{Current Practice in Diabetes Prevention}
\label{sec:current_practice_in_diabetes_prevention}

In clinical practice, preventive care for diabetes is allocated as follows \citep{Breeze.2017}: First, a health practitioner determines the patient's risk profile $r_i$ and then calculates a risk scoring to assess her risk of developing diabetes. In practice, the risk score is computed using a simple charting tool (\eg, the Framingham diabetes risk score). Such charting tools have an obvious limitation in that only a few variables from the risk profile are considered; a holistic assessment with complete EHRs is absent. Afterward, the patient is enrolled in preventive care if the risk score exceeds a certain threshold $\psi$, \ie,
\begin{equation}
\SingleSpacedXI
    t_i =   \begin{cases}
            1, & \text{if} \quad r_i \geq \psi , \\
            0, & \text{otherwise} .
            \end{cases}
\end{equation}%
The threshold $\psi$ is typically determined by experts in the field. To this end, we compute the threshold $\psi$ for enrolling patients in preventive care, such that the number of patients enrolled is set to $k$ per year---that is, the $k$ patients with the highest risk scores are eligible for the preventive care program. 

In our evaluation, the Framingham diabetes risk score typically is used to assess diabetes risk. The Framingham diabetes risk score was developed by the \US National Institutes of Health \citep{Wilson.2007} and represents the quasi-standard in clinical practice \citep{Long.2016}. The score is calculated using the charting tool shown in \Cref{tbl:framingham_scores}. Note that only a few risk factors are considered (\ie, fasting glucose level, BMI, cholesterol, parental history of diabetes mellitus, triglycerides, and blood pressure); all other variables from a patient's EHR are ignored. The failure to account for other variables thereby points toward opportunities for more effective risk assessment and prevention.

\begin{table}
	\TABLE
	{Calculation of the Framingham diabetes risk score.\label{tbl:framingham_scores}}
	{\scriptsize
	\begin{tabular}{l r}
			\toprule
			\textbf{Risk factor} & Score \\ 
			\midrule
			If fasting glucose level between 100 and 126mg/dL & +10 \\
			If body mass index between 25.0 and 29.9 & +2 \\
			If body mass index $\geq$ 30.0 & +5 \\
			If cholesterol level $<$ 40 mg/dL in men or $<$ +50 mg/dL in women & +5 \\
			If parental history of diabetes mellitus & +3 \\
			If triglyceride level $\geq$ 150 mg/dL & +3 \\
			If blood pressure $\geq$ 130/85 mm Hg or receiving treatment & +2 \\
			\midrule
			Framingham diabetes risk score $r_i$ & \text{$\sum$} \\
			\bottomrule
	\end{tabular}}
	{}
\end{table}

\subsection{Performance Metrics}
\label{sec:metrics}

We measure the operational performance of preventive care in such a way that the measure captures current practice at our IHO. Recall that preventive care aims to maximize the success of its preventive efforts \citep{WHO.2018}. In other words, given a certain budget, the expected number of disease onsets should be minimized. In medical decision-making, this measure is the quasi-standard in which outcome optimization is prioritized over cost minimization alone because of ethical and legal principles \citep{NHS}. In accordance with this standard, the expected number of prevented onsets represents our primary evaluation metric. Because health management also is interested in the economic value of implementing preventive care programs, we further report the projected cost savings but emphasize their secondary role.

We evaluate the performance of our decision model in a dynamic setting. Formally, we apply our decision model over a horizon of $l=1,\ldots,L$ time steps (here, years 2003 through 2012). Each year allows access only to historical data to assign patients to preventive care. We denote that patient $i$ was treated in year $l$ by $t_{i,l} = 1$. The preventive treatment of patient $i$ reduces the risk of developing diabetes in the future---that is, the treatment effect is $\gamma_{i,l}$. However, the outcomes of preventive treatment are not observable until months after the treatment. Let $y_{i,l+1} = 1$ denote the presence of diabetes for patient $i$ in the following year, whereas $y_{i,l+1} = 0$ describes the absence of diabetes. To be realistic, we consider censoring analogously to how it occurs in practice. On the one hand, we account for patients who have died and exclude them from being targeted for preventive treatment. On the other hand, we do not discount patients where a follow-up was missed (and also do not exclude them from being targeted for a preventive treatment) because these patients can incur a cost in subsequent years. To this end, our performance metrics are as follows:

\textbf{Prevented onsets.} We evaluate our decision model by computing the expected number of prevented onsets, which directly corresponds to the objective of our decision model and of preventive care in practice. Hence, it is our prime metric of interest. The expected number of prevented onsets in the patient cohort amounts to
\begin{equation}
\mathit{PreventedOnsets} = \frac{1}{L} \, \sum_{l=1}^L \, \sum_{i=1}^{N} \, \gamma_{i,l} \, t_{i,l} \, y_{i,l+1} ,
\end{equation}
where $\gamma_{i,l} \, t_{i,l} \, y_{i,l+1}$ gives the expected reduction in the risk of a diabetes onset for patient $i$ in year $l$, which is then averaged over the complete study horizon $1,\ldots,L$. 

\textbf{Cost savings.} We calculate the expected cost savings of our allocation for our IHO as follows. The cost savings depend on the cost of enrolling patients in preventive care and the cost of treating diabetes. A preventive treatment (here: \emph{metformin}) incurs an annual cost of $C_\text{prevent} = \$1,380$ per patient \citep{Shuyan.2015}. After a potential onset, the annual cost of treating diabetes depends on the age of the patient, as well as additional comorbid conditions---that is, $C_\text{diab} = C_0 + C_1 + \ldots + C_5$, where $C_0$ describes base costs that depend on the age of the patient, and $C_1, \ldots, C_5$ describe costs that can arise from comorbid conditions. In our case, we use $C_0 = \frac{\$15,000}{1 + \exp(-\mathit{age}/10)}$ for base costs and consider five common comorbid conditions of diabetes: acute myocardial infarction ($C_1 = \$5,000$), intracerebral hemorrhage ($C_2 = \$5,000$), acquired hypothyroidism ($C_3 = \$5,000$), angina pectoris ($C_4 = \$15,000$), and heart failure ($C_5 = \$15,000$).

We then calculate the expected cost for patient $i$ in year $l$ when being treated using
\begin{equation}
C_{i,l}^{\mathrm{(t)}} = y_{i,l+1} \, (1 - \gamma_{i,l}) \, C_\text{diab} + C_\text{prevent},
\end{equation}
and when not being treated using
\begin{equation}
C_{i,l}^{\mathrm{(nt)}} = y_{i,l+1} \, C_\text{diab}.
\end{equation}
By summing over the patients and the study horizon, we compute the cost savings as
\begin{equation}
\label{equ:cost}
\begin{split}
\mathit{CostSavings} = & \underbrace{\sum_{l=1}^{L} \, \sum_{i=1}^N C_{i,l}^{\mathrm{(nt)}}}_{\substack{\text{Cost for no}\\ \text{preventive treatments}}} \, - \, \underbrace{\sum_{l=1}^{L} \, \sum_{i=1}^N t_{i,l} \, C_{i,l}^{\mathrm{(t)}} + (1 - t_{i,l}) \, C_{i,l}^{\mathrm{(nt)}}}_{\substack{\text{Cost when allocating}\\ \text{patients to preventive program}}}.
\end{split}
\end{equation}
Of note, some costs are ongoing and thus are incurred beyond the study horizon covered by our dataset. To account for this extension, we calculate the expected costs using an expected lifetime of 75 \citep{tachkov2020life}. For older patients, we take into account a minimum of three additional years of life, and for younger patients, we consider a maximum of ten years. In doing so, we reflect that the longer remaining lifetime of younger patients means that diabetes is likely to lead to larger overall costs for treatment. 

\section{Model Development}
\label{sec:model_development}

\subsection{Problem Formulation}
\label{sec:problem_statement}


The objective of our partnering IHO is to dynamically allocate preventive care to patients so as to maximize the number of prevented cases of diabetes mellitus, given the available budget. Let $i=1, \ldots, N_+$ refer to the customers of the IHO---that is, to the patients in our dataset. The plus symbol indicates that some of the patients have already received preventive treatments, whereas the majority has not been treated. Let the first $N < N_+$ customers denote patients who have not been treated yet and therefore potentially benefit from being enrolled in the preventive care program. Simply enrolling all patients in preventive care is not economically feasible because of budget constraints. Instead, in each year $l=1,\ldots,L$, a decision must be made as to whether preventive care is warranted to patient $i$ ($t_{i,l} = 1$) or is not warranted ($t_{i,l} = 0$). 


For the purpose of decision-making, our IHO has access to additional information on patients' risk profiles through their EHR. EHR systems, similar to the one providing data for our evaluations, are used worldwide, thus ensuring the broad applicability of our model. For simplicity, we refer to the variables in the EHR by $x_{i,l} \in \mathbb{R}^n$, $i = 1, \ldots, N_+$, $l = 1, \ldots, L$. These variables encode the heterogeneity among patients and are used as predictors in our data-driven approach. Specifically, we use the EHRs for identifying patients at risk of developing diabetes, which is relevant in clinical practice because only a small subset of prediabetic patients actually develop diabetes. Hence, by accurately identifying patients who will experience an onset, preventive treatments can be allocated to patients who would most benefit from them. 


The objective is to maximize the expected effect of the preventive treatments. In other words, the expected number of disease onsets should be minimized within our time frame. As such, the decision of whether to enroll a specific patient in preventive care is driven only by the expected risk reduction of the preventive treatment for the patient \citep{Paulweber.2010}. IHOs also might have to consider additional budget restrictions---not at the patient level, but at the population level. Hence, the decision problem is given by  
\begin{mini!}
  {\substack{t_{1,1}, \ldots, t_{1,L}\\\ldots\\t_{N,1}, \ldots, t_{N,L}}}{\sum_{l=1}^L \left( \sum_{i=1}^{N} t_{i,l} \, \mathbb{E}\left[ y \, | \, x_{i,l}, t_{i,l} = 1\right] + \sum_{i=1}^{N} (1 - t_{i,l}) \, \mathbb{E}\left[ y \, | \, x_{i,l}, t_{i,l} = 0 \right] \right)\label{equ:objective}}{\label{def:decision_making_problem}}{}
\addConstraint{\sum_{i=1}^N t_{i,l}\label{equ:constr1}}{ \leq k, \quad \forall l \in \{1, \ldots, L\}}
\addConstraint{t_{i,l}\label{equ:constr2}}{\in \{0,1\}, \quad \forall i \in \{1, \ldots, N\} \forall l \in \{1, \ldots, L\}.}
\end{mini!}
\Cref{equ:objective} encodes the objective whereby the expected number of disease onsets is minimized. Here, the expectation $\mathbb{E}\left[ y \, | \, x_{i,l}, t_{i,l} = 1 \right]$ denotes the probability of developing diabetes, conditional on the EHR of patient $i$ in year $l$ and in the presence of preventive treatment ($t_{i,l} = 1$). The expectation $\mathbb{E}\left[ y \, | \, x_{i,l}, t_{i,l} = 0 \right]$ denotes the probability of an onset, conditional on the EHR of patient $i$ in absence of a preventive treatment ($t_{i,l} = 0$). Neither expectation is known, implying that the treatment effect must be estimated from data---for example, from electronic health records or, alternatively, by conducting randomized controlled trials. \Cref{equ:constr1} introduces a budget constraint on the number of patients that can be enrolled in preventive care, thus allowing the treatment of $k$ patients per year. \Cref{equ:constr2} makes this problem an integer problem. 

Our decision problem focuses on the expected number of diabetes onsets, similar to clinical practice; therefore, the cost of preventive treatments is not considered. Nevertheless, we can show that minimizing onsets also minimizes the overall cost of treatment (comprising both preventive treatments and treatments after a possible onset of the disease), given mild assumptions. This mathematical property is of direct relevance for health management because it underscores that improving the effectiveness of preventive care through better allocations also has financial value for healthcare organizations.

\begin{proposition}[Link of prevented diseases to cost minimization]
\mbox{Let $t_{1,l},\ldots,t_{1,L}, \ldots, t_{N,1},\ldots,t_{N,L}$} be a solution to the decision problem in \Cref{def:decision_making_problem}, let $C_\text{\textup{prevent}}$ denote the costs for prescribing preventive treatments, and let $C_\text{\textup{diab}}$ denote the costs for treating the disease after onset. Then, if 
\begin{equation}
t_{i,l}\,\left(C_\text{\textup{diab}} \, \mathbb{E}\left[ y \, | \, x_{i,l}, t_{i,l} = 1 \right] + C_\text{\textup{prevent}}\right) \leq t_{i,l}\,\left(C_\text{\textup{diab}} \, \mathbb{E}\left[ y \, | \, x_{i,l}, t_{i,l} = 0 \right]\right), \quad \forall i \in \{1,\dots,N\} \forall l \in \{1, \ldots, L\},
\end{equation} 
it follows that the solution $t_{1,l},\ldots,t_{1,L}, \ldots, t_{N,1},\ldots,t_{N,L}$ also minimizes costs for the health insurer.
\label{prop:link_to_costs}
\end{proposition}
\proof{Proof}
See Appendix~\ref{proof:link_to_costs}.
\endproof
\noindent

\subsection{Proposed Decision Model}
\label{sec:proposed_decision_model}

\begin{figure}[H]
\FIGURE
{\includegraphics[width=1.0\textwidth]{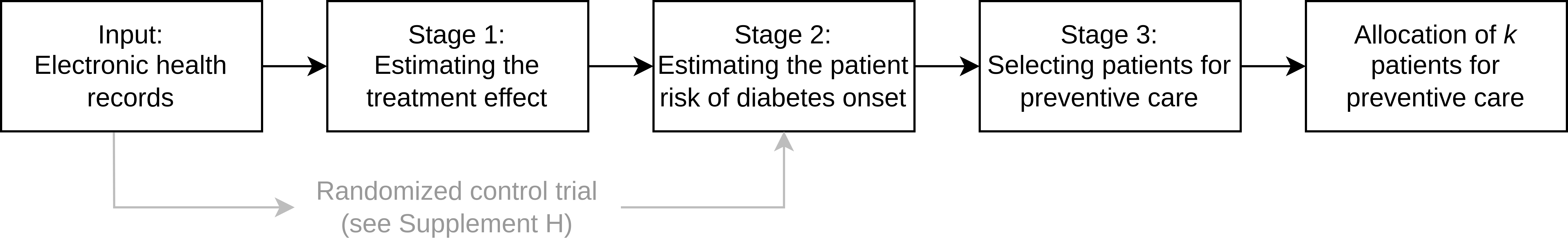}}
{Decision model for data-driven allocations of preventive care.\label{fig:framework}}
{}
\end{figure}

We develop a data-driven decision model for allocating preventive care---that is, a decision model that receives EHRs from patients as input and for which the output is a cost-effective allocation of preventive care resources (\Cref{fig:framework}). Our decision model uses a combination of causal inference, machine learning, and optimization, structured in three consecutive stages: (1)~The first stage uses counterfactual inference to estimate the effectiveness of preventive treatments using EHRs (\ie, the heterogeneous treatment effect). Thus, we quantify how the personalized diabetes risk of a patient with a specific risk profile changes when a preventive treatment is prescribed. Here, we serve needs in practice, where the treatment effectiveness is often unknown but must be estimated from EHRs. Later, we also discuss an alternative approach that can be used when the treatment effect is known (\eg, from a randomized controlled trial). (2)~The second stage estimates the personalized risk of a diabetes onset for an individual patient. Specifically, we apply machine learning to a patient's EHR to predict the probability of an onset. (3)~The third stage uses the predicted risk together with the treatment effect to optimize the allocation of preventive care to patients under the given budget constraint.

\subsubsection{Stage 1: Estimating the Treatment Effect.} 

In the first stage, we estimate the heterogeneous treatment effect of \emph{metformin}---that is, the expected reduction in the probability of a diabetes onset when prescribing \emph{metformin} to patients, given a specific patient risk profile. To account for heterogeneity in the treatment effectiveness across different patients, we introduce the following notation: Let $\{ 1, \ldots, N_+ \}$ be the set of patients in our sample, including patients who already received preventive treatment through \emph{metformin}, as well as patients who have not received preventive treatments. Let $I = \{ (i,l) \,: \, i\in\{1,\ldots,N_+\}, l\in\{1,\ldots,L\}\}$ be the set of 2-tuples for indexing patient-level observations throughout our study period. 


We then estimate the treatment effect for each patient $i$ in year $l$ via
\begin{equation}
    \gamma_{i,l} = \mathbb{E}\left[ y \, | \, x_{i,l}, t_{i,l} = 0 \right] - \mathbb{E}\left[ y \, | \, x_{i,l}, t_{i,l} = 1 \right].
\end{equation}
As a result, we allow the treatment effect to vary across patients who have different risk profiles. Building on the potential outcomes framework, it can further be shown that the treatment effects are identifiable from observational data \citep{Rubin.2005}.


To estimate the heterogeneous treatment effects, we use an established method: the causal forest \citep{Wager.2018}. The causal forest is a non-parametric method that extends the widely used random forest algorithm by \citet{breiman2001random} for estimating heterogeneous treatment effects. This approach comes with many benefits. First, the causal forest is an ensemble method, which means it combines the predictions of multiple simpler models to improve its performance. As a result, the causal forest works well in settings where the treatment effect is subject to complex, nonlinear relationships. Second, under common mathematical assumptions, the causal forest has been shown to be pointwise consistent for the true treatment effect and thus leads to provably valid inferences. Third, the causal forest is capable of handling high-dimensional and mixed data types with little risk of overfitting, making it a robust choice that requires minimal preprocessing.  Mathematically, one of these characteristics is its honesty property, where the trees in the causal forest are grown using one subsample of the dataset, while the predictions at the leaves of the tree are estimated using a different subsample \citep{Wager.2018}. By ensuring that each tree is only fit to a portion of the data, the honesty property helps to mitigate the risk of overfitting and to produce accurate and reliable treatment effect estimates. We report details on our implementation in \Cref{appendix:tuning_parameter}.

As part of our robustness checks, we perform a causal sensitivity analysis \citep{cinelli2020making} in which we study the effect of potential unobserved confounding variables on the estimated treatment effects. We find that, for unobserved confounding variables that are even strongly associated with our treatment, the treatment effects remain robust (see Supplement~\ref{appendix:robustness_to_unobserved_confounders}). Furthermore, we study how errors in the treatment effect estimation influence the overall performance of our decision model, and we find that the results remain largely robust. Finally, we provide an extension to settings where the treatment effect is not estimated from historical data but is given ex ante (\eg, where treatment effect is obtained through a randomized controlled trial instead), and we show how our decision model is applicable to such cases (see Supplement~\ref{appendix:personalized_coaching}).

\subsubsection{Stage 2: Estimating the Patient Risk of Diabetes Onset.} 

In the second stage, we rely on machine learning to estimate the heterogeneous risk of diabetes onset (in the absence of preventive treatment). For this purpose, we first filter out patients who already received preventive treatment. Let the first $N < N_+$ customers be patients who have not already been enrolled in a preventive care program. We then train a machine learning model $h_{0}^{\theta}$ with parameters $\theta \in \Theta$ as follows. The machine learning model uses EHRs, $x_{i,l}$, to predict the risk of the individuals' developing the disease. Formally, this is given by $\mathbb{E}\left[ y \, | \, x_{i,l}, t_{i,l} = 0 \right]$, $i \in \{1,\dots, N\}$, $l \in \{1,\ldots, L\}$. Hence, we have $h_{0}^{\theta}(x_{i,l}) = 1$ if there is an onset in the following year, and $h_{0}^{\theta}(x_{i,l}) = 0$ if the condition for patient $i$ is absent in the following year. Given $h_{0}^{\theta}$ and the estimated treatment effect $\gamma_{i,l}$ from stage 1, we can then compute the probability of onset, conditional on a preventive treatment. Then, the risk of a diabetes onset under preventive treatment amounts to $(1 - \gamma_{i,l}) \, h_{0}^{\theta}(x_{i,l})$.


Mathematically, $h_{0}^{\theta}$ is estimated as follows. Let $\theta \in \Theta$ refer to a set of parameters in the machine learning model $h_{0}^{\theta}$. For estimation, we minimize the empirical risk; that is,
\begin{align}
h_{0}^* = \argmin \limits_{\theta \in \Theta} \sum_{l=1}^L\,\sum_{i=1}^N \mathcal{L}(h_{0}^{{\theta}}(x_{i,l}), y_{i,l+1}),
\end{align}
where $\mathcal{L}$ denotes a loss function that measures the error of $h_{0}^{\theta}$ in forecasting the onset of a disease. We further calibrate the output probabilities of $h_{0}^{\theta}$ using Platt scaling (see Supplement~\ref{appendix:model_calibration}).

We later report results from our decision model using gradient-boosted decision trees. Gradient-boosted decision trees \citep{friedman2001greedy} belong to the category of ensemble methods, which are known to perform well on complex datasets and have been used in other operational applications \citep[\eg,][]{Glaeser.2019}. Nevertheless, we also perform an extensive series of robustness checks, repeating the analysis with other machine learning methods (\eg, lasso, ridge regression, random forest, and deep neural network). For all machine learning models, hyperparameters were tuned using 10-fold cross-validation. The tuning procedure is reported in Supplement~\ref{appendix:tuning_parameter}.

\subsubsection{Stage 3: Selecting Patients for Preventive Care.} 

To accommodate budgetary constraints, we model the maximum number of preventive treatments as $k$ patients per year. Our objective is to identify a cost-effective allocation that minimizes the expected number of onsets of the disease. Hence, we formulate our optimization problem as
\begin{mini!}
  {\substack{t_{1,1},\ldots,t_{1,L}\\\ldots\\ t_{N,1},\ldots,t_{N,L}}}{ \sum_{l=1}^L \left( \sum_{i=1}^{N} t_{i,l} \, \left(1 - \gamma_{i,l}\right) \, h^*_{0}(x_{i,l}) + \sum_{i=1}^{N} (1 - t_{i,l}) \, h^*_{0}(x_{i,l}) \right)}{}{}
\addConstraint{\sum_{i=1}^{N} t_{i,l}}{ \leq k, \quad \forall l \in \{1,  \ldots, L\}}
\addConstraint{t_{i,l}}{\in \{0,1\}, \quad \forall i \in \{1,  \ldots, N\} \forall l \in \{1,  \ldots, L\}.}
\end{mini!}
This optimization problem extends \Crefrange{equ:objective}{equ:constr2}, in that we replace the (unknown) expected values with their corresponding predictions from the machine learning methods.

In the case of \emph{metformin}, the solution to the previous optimization problem is the subset of $k$ patients per year for which the estimated reduction in expected onset is higher than the reduction for patients not in the subset. We show this property in the following proposition.

\begin{proposition}[Optimal allocation]
Let $h^*_{0}$ be an oracle for the risk of an onset, and let $\gamma_{i,l}$ be the true heterogeneous treatment effect of patient $i$ in year $l$. The allocation that minimizes the expected number of onsets in the population $\{1,\ldots,N\}$ provides preventive treatments in year $l$ to patients $\{1,\dots,k\}$, where
\begin{equation}
 (1 - \gamma_{i,l}) \, h^*_{0}(x_{i,l}) \geq (1 - \gamma_{j,l}) \, h^*_{0}(x_{j,l}), \qquad \forall i \in \{1,\dots,k\} \, \forall j \in \{k+1,\dots,N\}.
\end{equation} 
\label{prop:optimal_allocation}
\end{proposition}
\proof{Proof}
See Supplement~\ref{appendix:proof_optimal_allocation}.
\endproof

\noindent
In our implementation, we follow best practices and assess uncertainty in our model using a bootstrap procedure \citep{Hastie.2009}. Specifically, we use 100 samples with replacement and then report both the mean and the standard deviation. 

\section{Empirical Results}
\label{sec:empirical_results}

\subsection{Performance of Data-Driven Allocation}
\label{sec:comp_data_driven_against_current_practice}

In this section, we evaluate our decision model and compare its performance against the clinical baseline, which we use to mimic healthcare practitioners who base their decision-making on the Framingham diabetes risk score (see \Cref{sec:current_practice_in_diabetes_prevention}). In addition, we report a na{\"i}ve baseline, in which the allocation of preventive treatments is not optimized but, instead, a subset of $k$ patients from the population is randomly chosen for each year. By comparing the na{\"i}ve baseline with our decision model, we can directly quantify the relative gain from stratifying the allocation to high-risk patients through a data-driven approach. The results are in \Cref{tbl:res_vs_current}. 


The na{\"i}ve baseline (\ie, random allocation) is estimated to prevent 489 onsets when treating $k$=10,000 patients per year. For the same budget allowing for $k$=10,000 preventive treatments, the clinical baseline is estimated to prevent 701 onsets. Meanwhile, our decision model would prevent 882 onsets---an improvement of more than 25 percent. The results for other budgets lead to similar conclusions. In sum, our decision model is considerably more effective than current practice in disease prevention. 


The use of our decision model offers considerable cost savings. A random allocation of preventive treatments leads to cost savings of \$6.319 million at $k$=1,000 and \$11.988 million at $k$=10,000. This outcome shows that preventive care can be economical, in that the cost for treating a diabetes onset is substantially larger than the cost for preventive treatments. The clinical baseline improves cost savings further: to \$7.340 million at $k$=1,000 and to \$15.350 million at $k$=10,000. In comparison, our decision model demonstrates consistent cost savings over both the na\"ive baseline and the clinical baseline across all budgets $k$. For $k$=1,000 and $k$=10,000, our IHO could generate additional savings, beyond that of the clinical baseline, of \$0.710 million and \$2.987 million, respectively. As a result, the cost savings in our decision model are substantially larger than those of the clinical baseline and, based on a comparison with $t$-tests, are statistically significant at common significance thresholds.

\begin{table}[H]
\TABLE
{Performance of our data-driven decision model.\label{tbl:res_vs_current}}
{\scriptsize
\begin{tabular}{l SSS SSS}
\toprule
& \multicolumn{2}{c}{$k$=1,000} & \multicolumn{2}{c}{$k$=5,000} & \multicolumn{2}{c}{$k$=10,000} \\
\cmidrule(l){2-3} \cmidrule(l){4-5} \cmidrule(l){6-7} 
& {Prevented} & {Cost} & {Prevented} & {Cost} & {Prevented} & {Cost} \\
& {diseases} & {savings} & {diseases} & {savings} & {diseases} & {savings} \\
\midrule
\textbf{Na{\"i}ve baseline}  &        53.438 &              6.319 &      250.422 &              8.889 &       489.331 &              11.988  \\
          &         (0.330) &              (0.235) &        (1.698) &              (0.258) &         (2.162) &               (0.249)  \\
\textbf{Clinical baseline}  &       134.798 &              7.340 &      446.948 &             11.996 &       701.942 &              15.350  \\
          &         (1.025) &              (0.245) &        (3.373) &              (0.258) &         (3.837) &               (0.265)  \\
\midrule
\textbf{Our decision model}  & 162.913 &              8.050 &      567.710 &             13.979 &       882.004 &              18.337 \\
          &         (0.902) &              (0.239) &        (2.612) &              (0.253) &         (3.934) &               (0.284) \\
\bottomrule
\multicolumn{7}{l}{Stated: mean performance (standard deviation in parentheses)}
\end{tabular}}
{\SingleSpacedXI\footnotesize{\textit{Note.} Performance metrics for allocating preventive care given a varying budget for enrolling $k$ patients per year into preventive treatments. Cost savings over no preventive care allocation are reported in USD millions.}}
\end{table} 

\subsection{Comparison of Patient Characteristics}

In this section, we provide a quantitative assessment of the difference between the allocations from the clinical baseline and from our decision model. For this work, we report descriptive statistics for patients who were enrolled in preventive care (see \Cref{tbl:descriptives_fram_ours}). The clinical baseline based on the Framingham diabetes risk score allocates preventive care to patients who, on average, are 41.36 years old, whereas our decision model allocates preventive care to patients who are, on average, 53.60 years old and thus substantially older. This increase is in line with medical research \citep{Knowler.2002}, which shows that a larger treatment effect from \emph{metformin} is observed in people above age 45. Here, we note that the clinical baseline based on the Framingham diabetes risk score does not consider age to be a predictor of diabetes onset \citep{Wilson.2007}, and for this reason, following current practice does not explicitly stratify the allocation of preventive care to older age groups. Furthermore, the clinical baseline based on the Framingham diabetes risk score selects patients with an average BMI of 31.31, while our decision model gives preference to patients with a larger BMI (average: 34.55). We discussed our observations with medical professionals specializing in diabetes care, who explained that the stratification from our decision model is beneficial because, in their experience, \emph{metformin} leads to reduction in body weight; they noted that such preventive efforts are especially effective when offered to overweight patients. In terms of gender, both allocations exhibit a similar distribution among men and women. 

\begin{table}
	\TABLE
	{Descriptive statistics of patients enrolled in preventive care for different allocations. \label{tbl:descriptives_fram_ours}}
	{\scriptsize
	\begin{tabular}{l SS SS SS}
		\toprule
		{Variable} & \multicolumn{2}{c}{{Overall population}} &
		\multicolumn{2}{c}{{Clinical baseline}} & \multicolumn{2}{c}{{Our decision model}} \\
		\cmidrule(lr){2-3}\cmidrule(lr){4-5}\cmidrule(lr){6-7}
		& {Mean} & {SD} & {Mean} & {SD} & {Mean} & {SD} \\ 
		\midrule
		Age (in years) & 47.07 & 17.93 & 41.36 & 7.69 & 53.60 & 27.03 \\
        Sex (0=male; 1=female) & 0.51 & 0.50 & 0.53 & 0.50 & 0.55 & 0.50 \\
        Systolic blood pressure (mm Hg) & 122.84 & 17.11 & 124.86 & 14.19 & 125.65 & 17.13\\
        Diastolic blood pressure (mm Hg) & 77.26 & 12.78 & 78.72 & 9.76 & 78.98 & 11.90\\
        Body mass index (kg/m$^2$) & 29.05 & 5.64 & 31.31 & 6.16 & 34.55 & 6.10  \\
        HbA1c (in \%) & 5.56 & 0.36 & 5.74 & 0.28 & 5.83 & 0.34 \\
		\bottomrule
		\multicolumn{3}{l}{SD = standard deviation; $k$=5,000} \\
	\end{tabular}}
	{}
\end{table}  

\subsection{Cost-Effectiveness Analysis}
\label{sec:cost_effectiveness}

In our cost-effectiveness analysis \citep[similar to][]{Ayvaci.2012}, we estimate the number of prevented disease onsets and cost savings across varying budgets---that is, for a varying number of patients being enrolled in preventive care.

\begin{figure}
\FIGURE
{\includegraphics[width=.6\textwidth]{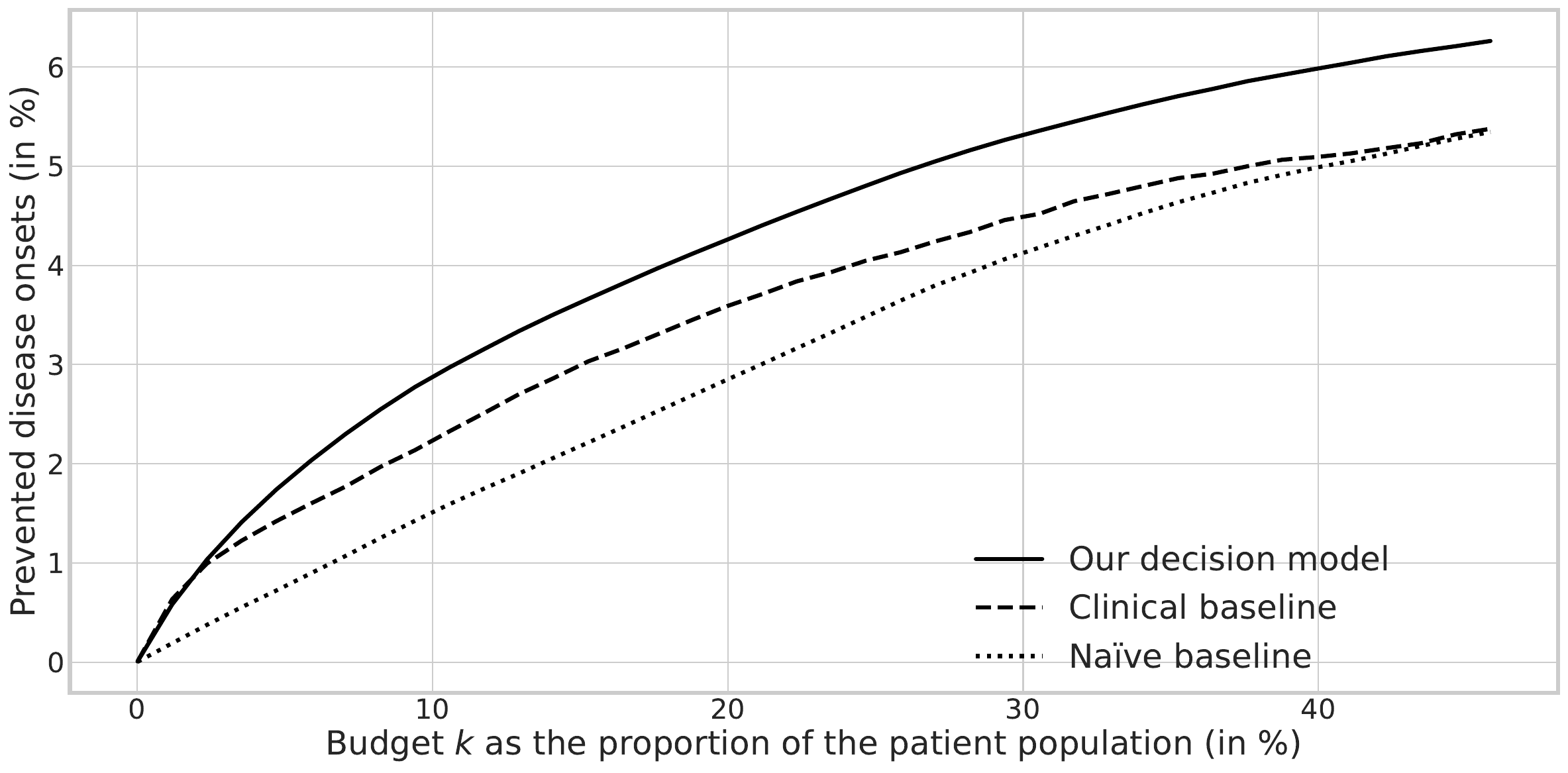}}
{Prevented onsets when providing preventive care under different budget constraints.\label{fig:disease_prev_varying_k}}
{}
\end{figure}

With increasing budgets, the relative proportion of prevented disease onsets also increases (see \Cref{fig:disease_prev_varying_k}). However, the improvement is more pronounced for our decision model, compared to both the na{\"i}ve baseline and the clinical baseline. Hence, for the same budget increase, our decision model is estimated to be more effective in preventing disease onsets. For instance, when enrolling \SI{10}{\percent} of the prediabetic population in preventive care, we estimate that the clinical baseline can prevent \SI{2.21}{\percent} of diabetes onsets. Using our decision model, this proportion increases to \SI{2.89}{\percent}. 

\begin{figure}
\FIGURE
{\includegraphics[width=.6\textwidth]{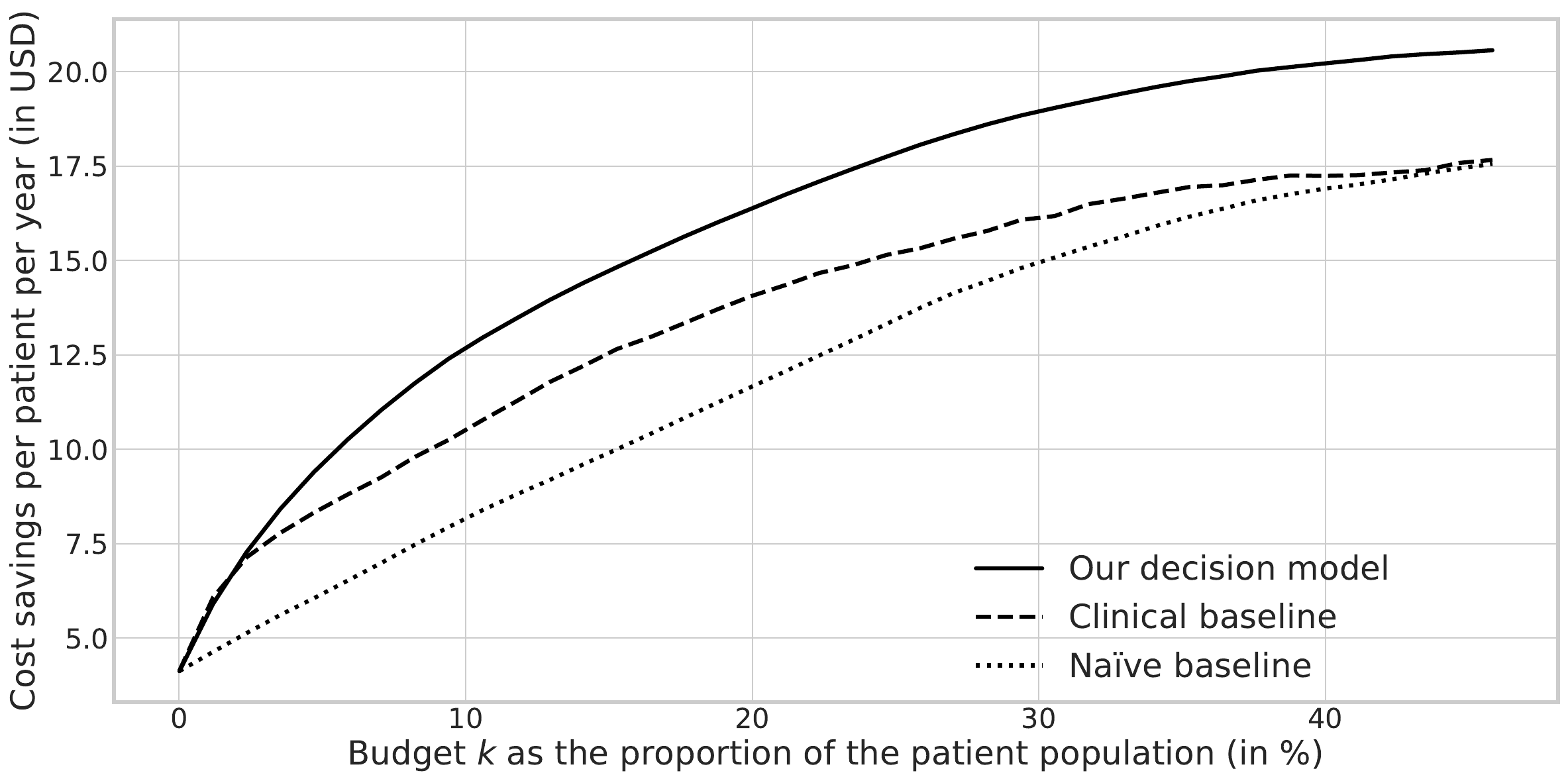}}
{Cost savings when providing preventive care under different budget constraints.\label{fig:costs_varying_k_per_patient}}
{}
\end{figure}

The cost savings over no preventive care are shown in \Cref{fig:costs_varying_k_per_patient}. As expected, the na{\"i}ve baseline and the clinical baseline based on the Framingham diabetes risk score achieve lower cost savings than our decision model, confirming again the superiority of our decision model. This finding is consistent across different budgets (\ie, different numbers of patients enrolled in preventive treatments). Our decision model leads to annual cost savings of up to \$20.31 per prediabetic patient. The cost savings per patient reach a maximum when \SI{47.58}{\percent} of the population is enrolled in preventive treatments. 

\subsection{Application to a \US-Wide Population}
\label{sec:summary_of_findings}


For better comparability, we now repeat the previous analysis for a population-wide setting, where we estimate the results of applying our decision model to the entire \US population. The \US population includes about 300 million adults, of whom 88 million are considered prediabetic. As previously noted, we assume that costs for preventive treatment are \$1,380 per patient and that model costs are dependent on age and potential comorbid conditions of a patient (see \Cref{sec:metrics}). We further assume the same machine learning performance in identifying high-risk patients as in the previous analyses.

On the basis of this setting, we obtain the following findings: Using the clinical baseline, allocating preventive treatments to \SI{10}{\percent} of patients (equal to 8.8 million people) leads to cost savings of \$913 million annually. This outcome corresponds to annual cost savings of \$10.38 per prediabetic patient. In comparison, our decision model offers larger benefits. When providing \SI{10}{\percent} of patients with preventive treatments, our decision model allows for annual cost savings of \$1.1 billion, equivalent to \$12.50 per prediabetic patient. Because the prevalence of diabetes in the United States is increasing \citep{Chen.2011}, the cost savings are likely to continue to grow in the coming years.

\subsection{Robustness Checks}

\subsubsection{Sensitivity analyses.}

We performed an extensive series of additional empirical analyses that add to the robustness of our findings: (1)~We tested our decision model with varying machine learning models (e.g., random forest, deep neural network). Details are in Supplement~\ref{appendix:choice_ml_model}. We find that all machine learning models perform well, and furthermore, the results favor our choice of gradient-boosted decision trees. (2)~We tested alternative risk scores (see Supplement~\ref{appendix:alternative_risk_scores}). Here, we find that our proposed decision model consistently leads to a larger number of prevented diseases and larger cost savings. (3)~We evaluated the robustness of the decision model to estimation errors in the treatment effect (see Supplement~\ref{appendix:robustness_to_estimation_errors}). We again find that our decision model shows robust results---in particular, when there is a larger number of patients in the treatment group. (4)~We demonstrate the applicability of our approach to personalized coaching as another form of preventive care (Supplement~\ref{appendix:personalized_coaching}).

\subsubsection{Explainability.}

We further build on explanation methods to analyze how our machine learning model arrives at predictions of diabetes onsets. To identify the most important predictors in our prediction model, we calculated SHapley Additive exPlanations (SHAP) values \citep{lundberg2017unified}. We find that the variables of age, BMI, and HbA1c are the most important predictors for our machine learning model when assessing the risk of a diabetes onset. Larger values for these variables result in the machine learning model's calculating a larger probability of diabetes onset. We discussed the results with medical professionals, who confirmed that our findings not only match their own experience but also are in line with medical research. Details are in Supplement~\ref{appendix:feature_importance}.

\subsubsection{Alternative Model Specifications.}
\label{sec:results_alternative_model_specifications}

We further benchmark different variations of our decision model to confirm the effectiveness of our model specification: 
\begin{quote}
\begin{enumerate}
    \item \emph{Main model.} We report the decision model from the main analysis. Here, the machine learning stage uses 307 input variables (see \Cref{tbl:variables}), and the optimization stage stratifies the allocation on the basis of risk reduction. 
    \item \emph{Sparse model I/II.} For Sparse Model I, we vary the input as follows: The model uses only a subset of input variables---namely, those from the Framingham diabetes risk score (see \Cref{tbl:framingham_scores}). For Sparse Model II, we also use the risk factors of age and the current glycated hemoglobin level (HbA1c). This addition allows us to quantify the performance gain from using modern, high-dimensional EHRs. 
    \item \emph{Linear model.} Here, we vary the machine learning stage by using only a linear model (with lasso regularization). By comparing this model against our main decision model (with gradient-boosted decision trees), we can assess the gains from modeling complex, nonlinear relationships (\eg, from comorbidities and other co-occurring conditions). 
    \item \emph{Risk-stratified model}. For the risk-stratified model, we vary the optimization stage, stratifying the allocation only by patient risk. That is, we select the patients for preventive care based on the highest risk of an onset, without considering heterogeneous treatment effects.
\end{enumerate}
\end{quote}

\noindent
\Cref{tbl:abl_study} presents the results for the five different model specifications. (1)~We find that the decision model from the main analysis performs best. (2)~We find that using high-dimensional patient data from modern EHRs is beneficial, but only to a relatively small degree. Replacing the 307 input variables with the variables from the Framingham diabetes risk score, together with age and HbA1c, can still lead to large improvements over the clinical baseline. In particular, the sparse model~II is ranked second overall, even though it has only seven input variables. Hence, the use of advanced machine learning, together with an optimization stage that accounts for heterogeneous treatment effects of preventive care, is more important than having access to rich EHR data. (3)~We find that nonlinear machine learning outperforms a linear model. The finding confirms the value of modeling complex, nonlinear relationships through nonlinear machine learning models. (4)~We obtain large benefits from including heterogeneous treatment effects in the objective. Hence, the optimization stage is more effective when allocation decisions take into account risk reduction, rather than only patient risk. 

\begin{table}[H]
\TABLE
{Results for different model specifications.\label{tbl:abl_study}}
{\scriptsize
\begin{tabular}{l SSS SSS}
\toprule
& \multicolumn{2}{c}{$k$=1,000} & \multicolumn{2}{c}{$k$=5,000} & \multicolumn{2}{c}{$k$=10,000} \\
\cmidrule(l){2-3} \cmidrule(l){4-5} \cmidrule(l){6-7} 
& {Prevented} & {Cost} & {Prevented} & {Cost} & {Prevented} & {Cost} \\
& {diseases} & {savings} & {diseases} & {savings} & {diseases} & {savings} \\
\midrule
\textbf{Clinical baseline}  &       134.798 &              7.340 &      446.948 &             11.996 &       701.942 &              15.350  \\
          &         (1.025) &              (0.245) &        (3.373) &              (0.258) &         (3.837) &               (0.265)  \\
\textbf{Main model}  & 162.913 &              8.050 &      567.710 &             13.979 &       882.004 &              18.337 \\
          &         (0.902) &              (0.239) &        (2.612) &              (0.253) &         (3.934) &               (0.284) \\
 \textbf{Sparse model I} & 137.209	&	6.745	&	466.106	&	11.364	&	728.151	&	15.041 \\
 & (0.757)	&	(0.195)	&	(2.152)	&	(0.210)	&	(3.326)	&	(0.234)	\\
 \textbf{Sparse model II}  &       159.148	&	7.850	&	549.777	&	13.492	&	855.743	&	17.752	\\
 & (0.880)	&	(0.231)	&	(2.532)	&	(0.246)	&	(3.848)	&	(0.276)	\\
\textbf{Linear model}  &  152.124	&	7.489	&	520.490	&	12.726	&	811.841	&	16.800	\\
& (0.840)	&	(0.218)	&	(2.401)	&	(0.234)	&	(3.684)	&	(0.261)	\\
\textbf{Risk-only model}  &        145.354	&	7.148	&	494.492	&	12.063	&	772.251	&	15.958	\\
& (0.802)	&	(0.207)	&	(2.282)	&	(0.222)	&	(3.523)	&	(0.249)	\\
\bottomrule
\multicolumn{7}{l}{Stated: mean performance (standard deviation in parentheses)}
\end{tabular}}
{\SingleSpacedXI\footnotesize{\textit{Note.} Performance metrics for allocating preventive care given a varying budget for enrolling $k$ patients per year into preventive treatments. Cost savings over no preventive care allocation are reported in USD millions.}}
\end{table} 

\section{Discussion}
\label{sec:discussion}

\subsection{Implications for Research}


Our novel decision model for data-driven allocation of preventive care integrates counterfactual inference, machine learning, and optimization. Specifically, the model allocates a given budget toward primary preventive care for patients at risk of developing a disease, so as to maximize the expected number of patients whose disease onset is prevented. We evaluate our decision model in a dynamic setting using a large, longitudinal dataset of 89,191 patients at risk of developing diabetes mellitus type~II, thereby confirming its effectiveness. 

We study the operational value of incorporating high-dimensional health data, machine learning, and optimization toward preventive treatment allocation decision-making. Through our experiments using a large, longitudinal dataset, we examined the performance gap between our proposed decision model and alternative model specifications (see \Cref{sec:results_alternative_model_specifications}). Overall, the largest performance gain in prevented diseases and cost savings comes from targeting patients based on risk reduction rather than only on risk. Hence, a crucial implication of our work is the importance of a rigorous modeling framework for decision-making. Also crucial is the use of advanced machine learning to accommodate complex, nonlinear relationships. A comparatively smaller gain comes for the use of high-dimensional EHRs. In fact, a decision model that uses machine learning and targets patients on the basis of risk reduction, but that has access only to a small subset of patient variables (sparse model~II), can even achieve a performance that is close to that of our main model. Our findings imply that using data-driven decision models to inform resource allocation in healthcare may be practical, even in cases where relatively little information is available about a patient. 

Proactively reducing individuals' risk of developing preventable diseases is a top priority in healthcare not only because it can improve patients' quality of life but also because it reduces healthcare costs. Our analysis shows that a large number of onsets can be prevented by using a data-driven decision model. In practice, allocation of patients to preventive care programs is often solely based on the risk of disease onset for the patients---that is, patients with the highest risk of onset are assigned to preventive treatments. However, prior medical literature has shown that preventive treatments can have a differential effectiveness \citep{Knowler.2002}. By exploiting this differential effectiveness, our method allocates preventive treatments in a more cost-effective manner, including prioritizing patients who have a lower risk of onset but who can benefit from preventive care more than patients who have a higher risk of onset. Similar observations have been made for targeting customers in marketing, where interventions can benefit when they consider not only the risk that a customer is about to churn, but also the incremental effectiveness of marketing interventions \citep{Ascarza2018}.

In our decision model, we also address the problem of unobserved confounders using a powerful statistical framework \citep{cinelli2020making}. Unobserved confounders simultaneously influence both the probability of being assigned to preventive treatment and the likelihood of developing diabetes. Thus, if unobserved confounders are not accounted for, the estimates of the treatment effect can be biased \citep{frauen2023sharp}. Such confounders are common in medicine and can include such variables as patients' socioeconomic status, healthcare coverage, and race. Often, these variables either cannot be directly observed or cannot be collected and used because of anti-discrimination laws. In light of this limitation, statistical frameworks that can address the potential effects of unobserved confounders can offer significant practical value in operations management.

\subsection{Implications for Health Management}

Healthcare managers have had to confront rising costs for care in recent years \citep{KC.2020}. To counteract these costs, regulatory initiatives, such as the Affordable Care Act (\S 4001(f)(3)), dictate a greater focus on disease prevention \citep{Koh.2010}. Health management research has long studied the efficient provision of care for patients \emph{after} they develop a medical condition \citep[\eg,][]{garg2020, zhang2016}; in contrast, our work charts a different path, where care is allocated \emph{prior} to any disease onset, with the goal of preventing it. To support health management in this task, we develop a novel data-driven decision model.


Current practice in allocating preventive care is based on risk scores and has several significant shortcomings. First, traditional risk scores (\eg, charting tools) are limited to a small subset of relevant patient variables that are commonly available in contemporary EHRs. Here, leveraging a few key variables from modern EHRs can offer better predictive power in identifying patients at risk of developing diabetes. Second, traditional risk scores also are limited to simplified linear models thus neglecting complex, nonlinear relationships between risk factors. Yet, such nonlinear relationships are particularly common in the presence of comorbidities and other co-occurring conditions. Third, traditional risk scores allocate preventive care according to a patient's risk level but without considering the heterogeneous treatment effects of preventive treatments across patients---that is, the differential effectiveness in reducing the risk of onset for a patient. Our work's contributions address these three key limitations. 

\subsection{Practical Considerations}

Our decision model can be readily developed and integrated into existing health management practices. Specifically, the approach we develop here is a general-purpose framework that can produce cost-effective intervention programs by healthcare systems that have access to EHRs; our approach can be applied to maximize prevention of any preventable disease for which preventive treatment exists, such as cardiovascular diseases or respiratory diseases, and for a given preventive care budget. As such, our approach is neither limited to a particular preventable disease nor does it rely on having access to a specific kind of information. In this work, we present empirical evaluations when using real healthcare data commonly recorded in practice by standard EHR systems worldwide. Therefore, our decision model is applicable to other IHOs in which similar standard EHR systems are used. 


In recent years, there has been a transition in healthcare from reactive to proactive care, underscoring the importance of preventive practices. Proactive care is a comprehensive approach that focuses on identifying and mitigating risk factors to prevent the onset of diseases, in contrast to only treating diseases and conditions once they arise. By proactively reducing individuals' risks of developing preventable diseases, health management can improve patient outcomes and reduce the long-term cost and resource burden on the healthcare system \citep{Zhou.2020}. This approach aligns with the objectives of IHOs in many countries, including several European countries, Israel, Singapore, and others. \citep{WHO.2016}. In the \US, some healthcare systems have goals that are consistent with cost-effective interventions to benefit the long-term well-being of patients; they include health maintenance organizations (e.g., the Kaiser group, with over 12 million members \citep[e.g.,][]{kaiser_group}) and Accountable Care Organizations, which merge patient data from different providers, with the goal of promoting cost-effective care. Generally, IHOs aim to improve coordination and efficiency among healthcare services and typically prioritize preventive care and patients' long-term well-being more broadly. By having access to comprehensive patient information from different health providers, IHOs can make informed decisions to optimize the allocation of resources for preventive care. Our decision model is directly applicable in such contexts and thus facilitates proactive care in two ways: (1) by providing a data-driven, cost-effective allocation of limited preventive care resources, and (2) by supporting health managers with the goal of improving patient outcomes and mitigating the burden of preventable debilitating diseases. Worldwide, evidence on the value of care and the use of data-driven care have led to an increasing trend toward integration in the healthcare sector \citep[e.g.,][]{WHO.2016, cuesta2019vertical}. In contrast, in countries where IHOs are not available, our decision model can motivate and promote collaboration among providers and insurers to enable cost-effective preventive care.

Operationally, the frequency of risk assessment and subsequent (re)allocation of a preventive treatment to patients ought to reflect the particular context. For instance, the onset risk of many preventable chronic diseases, including metabolic diseases such as diabetes mellitus type II, can vary over time, and a periodic estimation of risks and allocation of treatments, as we have done in this work, would be prudent. We have demonstrated how annual reallocation of preventive treatments can meaningfully reduce the number of patients affected by diabetes mellitus. In other contexts, the updating frequency ought to reflect the rate at which patients' risk is likely to change. As we have demonstrated here, our decision model can be applied to yield updated risk updates and to reallocate the preventive treatments accordingly.

Primary prevention considers all patients and prioritizes offering intervention treatments to patients based on their respective expected risk reduction. Thus, preventive care can yield the greatest benefits when it is not limited to considering only a subset of patients who initiate a physician's visit; instead, it could be proactively offered to patients for whom preventive care can be most effective, including patients who may be unaware of their own risk and thus do not seek treatment. In particular, given that nine in ten \US adults are unaware of their high-risk status \citep{Geiss.2010}, proactive efforts are paramount for effective diabetes prevention.

Implementing a decision model for the allocation of preventive care is subject to several practice challenges. One challenge is to ensure the availability of EHRs, which is a crucial requirement for identifying patients at risk of disease onset and for estimating heterogeneous treatment effects. Another challenge is the quality of EHR data, especially with respect to completeness and accuracy, both of which can affect the performance of the decision model. Finally, there may be regulatory challenges in implementing the decision model, such as privacy and data protection regulations that restrict access to patient data. As in any data-driven approach, establishing that the use of information to allocate disease prevention is consistent with local privacy and regulatory frameworks is important. Future work may also extend our analysis through the use of algorithmic fairness \citep{de2022algorithmic}.

\subsection{Concluding Remarks}
\label{sec:conclusion}


In an effort to reduce costs and improve quality of life, health management places an increasing focus on disease prevention. However, healthcare organizations lack the decision support necessary to facilitate effective allocation of preventive treatments to patients, given that population-wide access of preventive care is often prohibitively costly. To aid health management in this task, we have developed in this work a data-driven decision model for the cost-effective allocation of preventive care.

\ACKNOWLEDGMENT{%
Stefan Feuerriegel acknowledges funding from the Swiss National Science Foundation (SNSF) via Grant 186932. Mathias Kraus acknowledges funding from the Federal Ministry of Education and Research (BMBF) on "White-Box-AI" (Grant 01IS22080).
}

%
%
%


{\sloppy
\setlength{\bibsep}{0pt plus 0.3ex}
\bibliographystyle{informs2014}
\bibliography{literature}
}


\clearpage
\begin{APPENDICES}

\setcounter{page}{1}

\vspace{1cm}
\begin{center}
\Large Online Supplement
\end{center}
\vspace{0.5cm}

\section{Link of Prevented Diseases to Cost Minimization}
\label{proof:link_to_costs}
\proof{Proof of \Cref{prop:link_to_costs}.}
Without loss of generality, we study year $l$ within our study period, and let $\{1,\ldots,k\} \subset \{1,\ldots,N\}$ denote the patients that were enrolled for preventive treatment according to our decision model, i.e., which describe a solution to the decision problem (\Cref{def:decision_making_problem}). For simplicity, we neglect the index denoting the year. Then, the first part of this proof shows that the solution $\{1,\ldots,k\}$ minimizes costs among all subsets containing $k$ patients. The second part proves that there is no subset with fewer patients ($\{1,\ldots,k-1\} \subset \{1,\ldots,N\}$) that yields lower costs. 

\textbf{1st Part.} Because $\{1,\ldots,k\} \subset \{1,\ldots,N\}$ is a solution to the decision problem, we can assume that 
\begin{equation}
\begin{gathered}
    \mathbb{E}\left[ y \, | \, x_i, t_i = 0 \right] - \mathbb{E}\left[ y \, | \, x_i, t_i = 1 \right]
    \geq \mathbb{E}\left[ y \, | \, x_j, t_j = 0 \right] - \mathbb{E}\left[ y \, | \, x_j, t_j = 1 \right] , \\
     \forall i \in \{1,\dots,k\} \, \forall j \in \{k+1,\dots,N\}.
\end{gathered}
\end{equation}

We take a random patient $i^* \in \{1,\dots,k\}$ and a random patient $j^* \in \{k+1,\dots,N\}$ and exchange the treatment (\ie, we provide preventive treatment to patient $j^*$ but not to patient $i^*$). We compute the expected costs $C_\text{after}$ after this exchange and compare it to the expected costs $C_\text{before}$ before the exchange. The computation yields the following:
\begin{align}
& C_\text{after} - C_\text{before} \\ 
=\;& C_\text{diab} \, \left( \sum_{\substack{{i=1}\\i\neq i^*}}^{k} \mathbb{E}\left[ y \, | \, x_i, t_i = 1\right] + \sum_{\substack{{j=k+1}\\j\neq j^*}}^{N} \mathbb{E}\left[ y \, | \, x_j, t_j = 0 \right] + \mathbb{E}\left[ y \, | \, x_{j^*}, t_{j^*} = 1\right] + \mathbb{E}\left[ y \, | \, x_{i^*}, t_{i^*} = 0\right] \right) \notag \\ &+ k \, C_\text{prevent}
- C_\text{diab} \, \left( \sum_{i=1}^{k} \mathbb{E}\left[ y \, | \, x_i, t_i = 1\right] + \sum_{j=k+1}^{N} \mathbb{E}\left[ y \, | \, x_j, t_j = 0 \right] \right) - k \, C_\text{prevent} \\
=\;& C_\text{diab} \, \left( \sum_{\substack{{i=1}\\i\neq i^*}}^{k} \mathbb{E}\left[ y \, | \, x_i, t_i = 1\right] + \sum_{\substack{{j=k+1}\\j\neq j^*}}^{N} \mathbb{E}\left[ y \, | \, x_j, t_j = 0 \right] + \mathbb{E}\left[ y \, | \, x_{j^*}, t_{j^*} = 1\right] + \mathbb{E}\left[ y \, | \, x_{i^*}, t_{i^*} = 0\right] \right. \notag \\ 
&- \left. \sum_{i=1}^{k} \mathbb{E}\left[ y \, | \, x_i, t_i = 1\right] - \sum_{j=k+1}^{N} \mathbb{E}\left[ y \, | \, x_j, t_j = 0 \right] \vphantom{\sum_{\substack{{i=1}\\i\neq i^*}}^{k} \mathbb{E}} \right) \\
=\;& C_\text{diab} \, \bigg( \mathbb{E}\left[ y \, | \, x_{i^*}, t_{i^*} = 0\right] - \mathbb{E}\left[ y \, | \, x_{i^*}, t_{i^*} = 1\right] + \mathbb{E}\left[ y \, | \, x_{j^*}, t_{j^*} = 1 \right] - \mathbb{E}\left[ y \, | \, x_{j^*}, t_{j^*} = 0 \right] \bigg) \\ 
\geq\; & 0
\end{align}

\textbf{2nd Part.} We now show that no subset comprising fewer patients ($\{1,\ldots,k-1\} \subset \{1,\ldots,N\}$) yields lower costs. For this computation, we assume that:
\begin{align}
C_\text{diab} \, \mathbb{E}\left[ y \, | \, x_i, t_i = 1 \right] + C_\text{prevent} \leq C_\text{diab} \, \mathbb{E}\left[ y \, | \, x_i, t_i = 0 \right] \qquad \forall i \in \{1,\dots,k\}.
\end{align}
That an exchange of patients between the treatment group and the non-treatment group leads to increased costs directly follows from part 1. Thus, without loss of generality, we consider subgroups $\{1,\ldots,k-1\}$ of our original solution $\{1,\ldots,k\}$ to the decision problem. We compute costs $C_{k-1}$ when providing preventive care to $k-1$ patients and compare $C_{k-1}$ to the original costs $C_{k}$:
\begin{align}
&C_{k-1} - C_k \\
=\;& C_\text{diab} \, \left( \sum_{i=1}^{k-1} \mathbb{E}\left[ y \, | \, x_i, t_i = 1\right] + \sum_{j=k}^{N} \mathbb{E}\left[ y \, | \, x_j, t_j = 0 \right] \right) + (k - 1) \, C_\text{prevent} \notag \\ &- C_\text{diab} \, \left( \sum_{i=1}^{k} \mathbb{E}\left[ y \, | \, x_i, t_i = 1\right] + \sum_{j=k+1}^{N} \mathbb{E}\left[ y \, | \, x_j, t_j = 0 \right] \right) - k \, C_\text{prevent} \\
=\;& C_\text{diab} \, \left( \sum_{i=1}^{k-1} \mathbb{E}\left[ y \, | \, x_i, t_i = 1\right] + \sum_{j=k}^{N} \mathbb{E}\left[ y \, | \, x_j, t_j = 0 \right] - \sum_{i=1}^{k} \mathbb{E}\left[ y \, | \, x_i, t_i = 1\right] - \sum_{j=k+1}^{N} \mathbb{E}\left[ y \, | \, x_j, t_j = 0 \right] \right) - C_\text{prevent} \\
=\;& C_\text{diab} \, \bigg( \mathbb{E}\left[ y \, | \, x_k, t_k = 0 \right] - \mathbb{E}\left[ y \, | \, x_k, t_k = 1\right] \bigg) - C_\text{prevent} \\
=\;& C_\text{diab} \, \bigg( \mathbb{E}\left[ y \, | \, x_k, t_k = 0 \right] \bigg) - C_\text{diab} \, \left( \mathbb{E}\left[ y \, | \, x_k, t_k = 1\right] \right) - C_\text{prevent} \\
\geq\;& 0
\end{align}
\hfill\Halmos
\endproof

\clearpage

\section{Proof for Optimal Allocation}
\label{appendix:proof_optimal_allocation}

\proof{Proof of \Cref{prop:optimal_allocation}.} 
Let $n_\text{onsets}$ denote the number of expected onsets. Assume $h_0^*(x_{i,l})$ to be a perfect model for $\mathbb{E}\left[ y \, | \, x_{i,l}, t_{i,l} = 0 \right]$ and assume $\gamma_{i,l}$ to be the true treatment effect for $i \in \{1,\dots,N\}$. Let $\{1,\dots,k\}$ be the subset of patients who are prescribed preventive care, and let $\{k+1,\dots,N\}$ be the subset of patients who are not prescribed preventive care in year $l$. Let this allocation follow our described approach to allocating patients; thus,
\begin{equation}
(1 - \gamma_{i,l}) \, h^*_{0}(x_{i,l}) \geq (1 - \gamma_{j,l}) \, h^*_{0}(x_{j,l}), \qquad \forall i \in \{1,\dots,k\} \, \forall j \in \{k+1,\dots,N\} ,
\end{equation}
which is equal to
\begin{equation}
h^*_{0}(x_{i,l}) + \gamma_{j,l} \, h^*_{0}(x_{j,l}) \geq h^*_{0}(x_{j,l}) +  \gamma_{i,l} \, h^*_{0}(x_{i,l}) , \qquad \forall i \in \{1,\dots,k\} \, \forall j \in \{k+1,\dots,N\}.
\end{equation}

In the following computation, we take a random patient $i^* \in \{1,\dots,k\}$ and a random patient $j^* \in \{k+1,\dots,N\}$ and exchange the treatment---that is, we provide preventive treatment to patient $j^*$ but not to patient $i^*$. From this exchange, we arrive at the following:

\begin{align}
    n_\text{onsets} &= 
    \sum_{\substack{{i=1}\\i\neq i^*}}^{k} \mathbb{E}\left[ y \, | \, x_{i,l}, t_{i,l} = 1\right] + \sum_{\substack{{j=k+1}\\j\neq j^*}}^{N} \mathbb{E}\left[ y \, | \, x_{j,l}, t_{j,l} = 0 \right] + \mathbb{E}\left[ y \, | \, x_{j^*,l^*}, t_{j^*,l^*} = 1\right] + \mathbb{E}\left[ y \, | \, x_{i^*,l^*}, t_{i^*,l^*} = 0\right]\\
    &= \sum_{\substack{{i=1}\\i\neq i^*}}^{k} \left(1 - \gamma_{i,l}\right) \, h^*_{0}(x_{i,l}) + \sum_{\substack{{j=k+1}\\j\neq j^*}}^{N} h^*_{0}(x_{j,l}) + \left(1 - \gamma_{j^*,l^*}\right) \, h^*_{0}(x_{j^*,l^*}) + h^*_{0}(x_{i^*,l^*})\\
    &\geq \sum_{\substack{{i=1}\\i\neq i^*}}^{k} \left(1 - \gamma_{i,l}\right) \, h_0^*(x_{i,l}) + \sum_{\substack{{j=k+1}\\j\neq j^*}}^{N} h_0^*(x_{j,l}) + \left(1 - \gamma_{i^*,l^*}\right) h_0^*(x_{i^*,l^*}) + \, h_0^*(x_{j^*,l^*}) \\
    &= \sum_{i=1}^{k} \left(1 - \gamma_{i,l}\right) \, h_0^*(x_{i,l}) + \sum_{j=k+1}^{N} h_0^*(x_{j,l}) \\
    &= \sum_{i=1}^{k} \mathbb{E}\left[ y \, | \, x_{i,l}, t_{i,l} = 1\right] + \sum_{j=k+1}^{N} \mathbb{E}\left[ y \, | \, x_{j,l}, t_{j,l} = 0 \right] 
\end{align}
Thus, exchanging the treatments of patients $i^*$ and $j^*$ leads to the same or a larger number of expected onsets. Consequently, our original allocation minimizes the number of expected onsets.
\hfill\Halmos
\endproof

\clearpage

\section{Robustness Checks}
\label{appendix:robustness_checks}

\subsection{Choice of Machine Learning Model}
\label{appendix:choice_ml_model}

For the analysis in the main paper, the machine learning in stage~2 of the decision model was set to gradient boosted decision trees \citep{friedman2001greedy}. Gradient boosted decision trees belong to the category of tree ensemble methods, which have been applied successfully in other operational applications \citep[\eg,][]{Glaeser.2019,Senoner.2021}. We now provide empirical evidence supporting this choice. For this purpose, we compare the performance of our decision model under different machine learning models in stage~2 while the rest of the decision model remains the same. Our comparison includes linear (lasso and ridge regression) and nonlinear (random forest and deep neural network) models. All estimation details are provided in Supplement~\ref{appendix:tuning_parameter}.

Results are shown in \Cref{tbl:res_models}. While the lasso performs well among the linear models, nonlinear machine learning models consistently outperform linear models. This may be because nonlinear models are better able to exploit complex patterns in the data, particularly with regard to between-patient heterogeneity. However, nonlinear models can also be prone to overfitting, which may explain why the deep neural network underperformed compared to the gradient boosted decision trees. Overall, the decision model using gradient boosted decision trees achieves the best performance.

\begin{table}[H]
\TABLE
{Performance comparison among different machine learning models.\label{tbl:res_models}}
{\SingleSpacedXI
\scriptsize
\begin{tabular}{l SSS SSS}
\toprule
& \multicolumn{2}{c}{$k$=1,000} & \multicolumn{2}{c}{$k$=5,000} & \multicolumn{2}{c}{$k$=10,000} \\
\cmidrule(l){2-3} \cmidrule(l){4-5} \cmidrule(l){6-7} 
& {Prevented} & {Cost} & {Prevented} & {Cost} & {Prevented} & {Cost} \\
& {diseases} & {savings} & {diseases} & {savings} & {diseases} & {savings} \\
\midrule
\textbf{Na{\"i}ve baseline}  &        53.438 &              6.319 &      250.422 &              8.889 &       489.331 &              11.988  \\
          &         (0.330) &              (0.235) &        (1.698) &              (0.258) &         (2.162) &               (0.249)  \\
\textbf{Lasso} & 152.124	&	7.489	&	520.490	&	12.726	&	811.841	&	16.800	\\
& (0.840)	&	(0.218)	&	(2.401)	&	(0.234)	&	(3.684)	&	(0.261)	\\
\textbf{Ridge regression}  &  150.495	&	7.409	&	514.812	&	12.586	&	803.021	&	16.617 \\
& (0.831)	&	(0.216)	&	(2.374)	&	(0.231)	&	(3.644)	&	(0.258)	\\
\textbf{Random forest}  &  160.270	&	7.892	&	548.875	&	13.425	&	855.942	&	17.717	\\
& (0.885)	&	(0.230)	&	(2.531)	&	(0.246)	&	(3.880)	&	(0.276)	\\
\textbf{Deep neural network}  &  155.636	&	7.670	&	535.133	&	13.109	&	833.792	&	17.276	\\
& (0.860)	&	(0.224)	&	(2.467)	&	(0.240)	&	(3.766)	&	(0.268)	\\
\midrule
\textbf{Gradient boosted}  & 162.913 &              8.050 &      567.710 &             13.979 &       882.004 &              18.337 \\
\textbf{decision trees}            &         (0.902) &              (0.239) &        (2.612) &              (0.253) &         (3.934) &               (0.284) \\
\bottomrule
\multicolumn{7}{l}{Stated: mean performance (standard deviation in parentheses)}
\end{tabular}}
{\SingleSpacedXI\footnotesize{\textit{Note.} Performance metrics for allocation of preventive care when using the respective machine learning model in stage~2 of our decision model. Budget constraints allow for the treatment of $k$ patients per year. Cost savings are reported in USD millions.}}
\end{table} 

\subsection{Comparison with Alternative Risk Scores}
\label{appendix:alternative_risk_scores}

The Framingham diabetes risk score represents the quasi-standard in clinical practice for assessing diabetes risk \citep{Long.2016}. Nevertheless, \Cref{tbl:other_risk_scores} shows the comparison between our decision model and other diabetes risk scores from clinical practice---namely, \citet{Lindstroem.2003}, \citet{Wilson.2007}, \citet{Kahn.2009}, and \citet{Rosella.2011}. Consistent with our previous findings, our proposed decision model outperforms all alternative risk scores.

\begin{table}[H]
\TABLE
{Performance comparison between our decision model and other risk scores.\label{tbl:other_risk_scores}}
{\scriptsize
\begin{tabular}{l SSS SSS}
\toprule
& \multicolumn{2}{c}{$k$=1,000} & \multicolumn{2}{c}{$k$=5,000} & \multicolumn{2}{c}{$k$=10,000} \\
\cmidrule(l){2-3} \cmidrule(l){4-5} \cmidrule(l){6-7} 
& {Prevented} & {Cost} & {Prevented} & {Cost} & {Prevented} & {Cost} \\
& {diseases} & {savings} & {diseases} & {savings} & {diseases} & {savings} \\
\midrule
\textbf{Na{\"i}ve baseline}  &        53.438 &              6.319 &      250.422 &              8.889 &       489.331 &              11.988  \\
          &         (0.330) &              (0.235) &        (1.698) &              (0.258) &         (2.162) &               (0.249)  \\
\textbf{\citet{Lindstroem.2003}}  &  128.810	&	6.323	&	434.431	&	10.562	&	679.740	&	14.015	\\
& (0.710)	&	(0.181)	&	(2.008)	&	(0.196)	&	(3.126)	&	(0.219)	\\
\textbf{\citet{Wilson.2007}}  &  121.424	&	5.979	&	415.914	&	10.173	&	648.571	&	13.425	\\
& (0.671)	&	(0.174)	&	(1.918)	&	(0.187)	&	(2.940)	&	(0.209)	\\
\textbf{\citet{Kahn.2009}}  &  119.542	&	5.879	&	406.948	&	9.930	&	635.441	&	13.133\\
& (0.660)	&	(0.170)	&	(1.878)	&	(0.183)	&	(2.897)	&	(0.205)	\\
\textbf{\citet{Rosella.2011}}  &  120.917	&	5.940	&	409.335	&	9.966	&	639.950	&	13.207	\\
& (0.667)	&	(0.171)	&	(1.891)	&	(0.184)	&	(2.932)	&	(0.206)	\\
\midrule
\textbf{Our decision model}  & 162.913 &              8.050 &      567.710 &             13.979 &       882.004 &              18.337 \\
            &         (0.902) &              (0.239) &        (2.612) &              (0.253) &         (3.934) &               (0.284) \\
\bottomrule
\multicolumn{7}{l}{Stated: mean performance (standard deviation in parentheses)}
\end{tabular}}
{\SingleSpacedXI\footnotesize{\textit{Note.} Performance metrics for allocation of preventive care with the respective decision model. Budget constraints allow for the treatment of $k$ patients per year. Cost savings over no preventive care allocation are reported in USD millions.}}
\end{table} 

\subsection{Robustness to Estimation Errors in the Treatment Effect}
\label{appendix:robustness_to_estimation_errors}

We now analyze the robustness of our decision model to potential errors in the estimation of the treatment effect. For this, we run the same analysis as before, but introduce additional Gaussian noise to the estimated treatment effect $\gamma_{i,l}$, \ie,
\begin{equation}
    \tilde{\gamma}_{i,l} = \gamma_{i,l} + \epsilon \qquad \text{with} \qquad \epsilon \sim \mathcal{N}(0,\,\sigma^2),
\end{equation}
where $\sigma$ denotes the level of noise. The latter thus represents the estimation error. We further know that the effectiveness of preventive care such as \emph{metformin} is bounded and, therefore, ensure the noisy treatment effect $\tilde{\gamma}_{i,l}$ to lie in the interval $[0,1]$ through clipping. 

\Cref{tbl:noisy_treatment_effect} lists the results for this analysis. Here, we compare different noise levels $\sigma \in \{ 0, 0.1, 0.5 \}$. Of note, a noise level of $\sigma = 0.5$ introduces a large estimation error, especially when considering that the effectiveness of preventive treatments should lie in the interval $[0, 1]$ for real-world clinical settings. As expected, we find a slightly more dominant role of estimation errors for small $k$, which, in light of the small set of patients allocated to preventive care, results in slightly reduced cost savings. This can be expected due to the small sample size. However, and more importantly, we find that the impact of estimation errors is overall fairly small: we still achieve a large number of prevented disease onsets as well as large cost savings over current practice. For example, even for noise of $\sigma = 0.1$, our decision model still outperforms current practice for $k=10,000$ and is only slightly outperformed by current practice for noise of $\sigma = 0.5$. 

\begin{table}[H]
\TABLE
{Robustness of our decision model to estimation errors in the treatment effect.\label{tbl:noisy_treatment_effect}}
{\footnotesize
\begin{tabular}{l SSS SSS}
\toprule
& \multicolumn{2}{c}{$k$=1,000} & \multicolumn{2}{c}{$k$=5,000} & \multicolumn{2}{c}{$k$=10,000} \\
\cmidrule(l){2-3} \cmidrule(l){4-5} \cmidrule(l){6-7} 
& {Prevented} & {Cost} & {Prevented} & {Cost} & {Prevented} & {Cost} \\
& {diseases} & {savings} & {diseases} & {savings} & {diseases} & {savings} \\
\midrule
\textbf{Clinical baseline}  &       134.798 &              7.340 &      446.948 &             11.996 &       701.942 &              15.350  \\
          &         (1.025) &              (0.245) &        (3.373) &              (0.258) &         (3.837) &               (0.265)  \\
$\boldsymbol{\sigma = 0}$  & 162.913 &              8.050 &      567.710 &             13.979 &       882.004 &              18.337 \\
            &         (0.902) &              (0.239) &        (2.612) &              (0.253) &         (3.934) &               (0.284) \\
$\boldsymbol{\sigma = 0.1}$  & 149.481	&	7.331	&	501.655	&	12.172	&	785.779	&	16.180	\\
& (0.823)	&	(0.209)	&	(2.320)	&	(0.227)	&	(3.630)	&	(0.253)	\\
$\boldsymbol{\sigma = 0.5}$  & 131.199	&	6.462	&	449.977	&	11.012	&	701.492	&	14.526	\\
& (0.725)	&	(0.189)	&	(2.075)	&	(0.202)	&	(3.176)	&	(0.226)	\\
\bottomrule
\multicolumn{7}{l}{Stated: mean performance (standard deviation in parentheses)}
\end{tabular}}
{\SingleSpacedXI\footnotesize{\textit{Note.} Budget constraints allow for the treatment of $k$ patients per year. Cost savings over no preventive care allocation are reported in USD millions.}}
\end{table} 

\subsection{Robustness to Number of Training Samples}
In machine learning, the size of the training set is known to affect the prediction performance, which, in turn, affects the operational performance. We thus study the convergence of our decision model against the optimal decision for an increasing number of observations. We use a static setting with observations $x \in \mathbb{R}^3$, for which the three components represent different patient variables (age, height, body mass index). We simulate these as follows:

{\SingleSpacedXI\begin{align}
    x &\sim \mathcal{N}(\mu,\,\Sigma) , \\
    \mu &= (50, 170, 27)^T , \\
    \Sigma &= \begin{bmatrix} 20 & 0 & 5 \\ 0 & 50 & 5 \\ 5 & 5 & 5 \end{bmatrix}
\end{align}}
The true but unobservable risk of an onset is given by
\begin{equation}
    y = \sigma\left(\frac{0.5 \, x_1 + 0.1 \, x_2 + 0.2 \, x_3}{100}\right),
\end{equation}
with $\sigma$ denoting the sigmoid function. We assume that we observe the actual outcomes only for $N_{\text{train}}$ patients. That is, for each patient $i$, we assume that we observe the realization of $Y$ given by

{\SingleSpacedXI\begin{align}
    y_i = \begin{cases}
    1, & \quad \text{if} \quad y + \epsilon \geq 0.7,\\
    0, & \quad \text{otherwise}, \\
    \end{cases} 
\qquad\text{with}\qquad
\epsilon \sim \mathcal{N}(0,\,1) .
\end{align}}
In the simulation, we further set the true treatment effect to $\gamma = 0.31$.

Our population comprises $N=$\,\num{100000} patients. The budget constraint allows us to enroll $k=$\,\num{10000} patients into preventive care. \Cref{fig:simulation} shows the number of prevented diseases for varying training sample sizes $N_{\text{train}}$. It confirms that our decision model converges to the optimal allocation. Moreover, we already yield a close-to-optimum performance for a relatively small number of training samples (\ie, fewer than 3,000).

\begin{figure}[H]
\FIGURE
{\includegraphics[width=.8\textwidth]{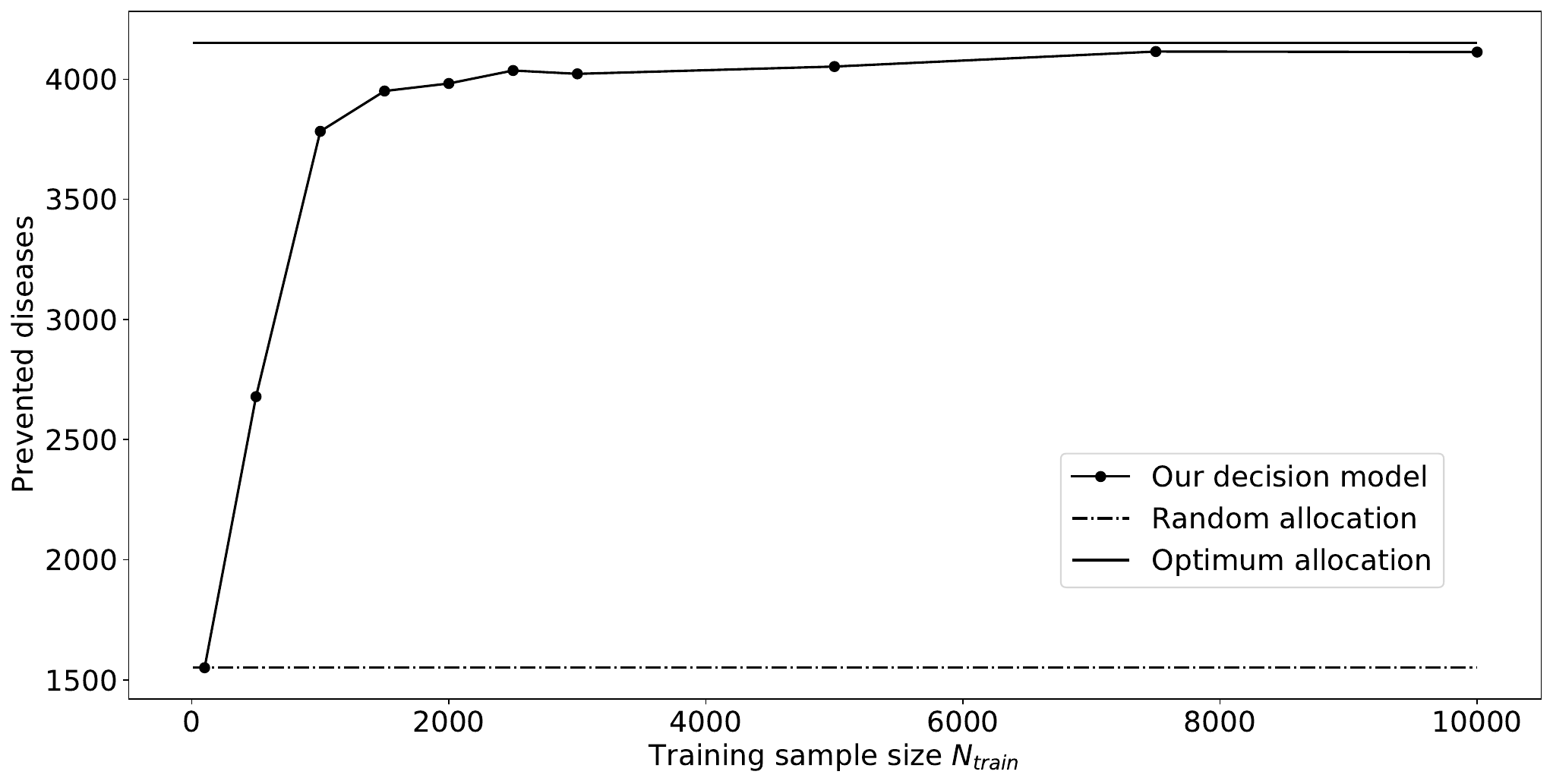}}
{Performance of our decision model for varying number of training samples.\label{fig:simulation}}
{}
\end{figure}

\clearpage

\section{Causal Sensitivity Analysis}
\label{appendix:robustness_to_unobserved_confounders}

In the following, we analyze how potential unobserved confounders would affect the estimated treatment effects. Unobserved confounders refer to unmeasured variables that affect both treatment assignment and the health outcome \citep{cinelli2020making}. A prominent example is the socioeconomic status of patients, as wealthier patients are often more likely to enroll in preventive care and, furthermore, can afford a healthier lifestyle that is responsible for a lower diabetes risk, which thus may underestimate the treatment effect of \emph{metformin}. Since these confounders are unobserved, we cannot control for them when estimating the effect of a treatment. As a remedy, we follow the so-called causal sensitivity analysis from \citet{cinelli2020making} and analyze the effect that an unobserved confounder would have on the treatment effect estimation, if it existed. More precisely, we select an observed variable which affects both treatment assignment and the health outcome and now assume that an unobserved confounder exists with the same strength as the observed one. Then, we check that the treatment effect cannot be explained away and thus remains robust. 

The age of a patient is generally associated with the likelihood of being prescribed with metformin \citep[e.g.,][]{Rosella.2011} and has also been shown to be strongly associated with the risk of diabetes onset \citep[e.g.,][]{Knowler.2002}. Thus, we use the strength of the variable \textquote{age} in the following as our reference to make comparisons of whether the treatment effect can be explained away. 

We follow \citet{cinelli2020making} and use a linear regression. Informed by clinical research (e.g., the Framingham risk score), we use a linear regression with the following patient variables, i.e.,
\begin{equation}
y = treatment + age + sex + weight + body\_mass\_index + systolic\_bp + diastolic\_bp + height.
\end{equation}
We then perform causal sensitivity analysis for all patients and first test for the sensitivity of the main effect. Afterward, to account for heterogeneity in the treatment effect, we repeat the causal sensitivity analysis for different patient subgroups. For this, we segment the patients by their estimated treatment effects into four groups -- A, B, C, D -- using a decision tree regressor, and then perform the causal sensitivity analysis for each subgroup. 

\Cref{tbl:unobs_conf} shows the results of our causal sensitivity analysis. The results confirm that the estimated treatment effect is robust to unobserved confounders. The results are also consistent across all subgroups. Note that the patient's age is considered to be one of the strongest predictors for diabetes \citep{american2022}, and, thus, it is very unlikely to have an unobserved confounder of a similar strength. In other words, the treatment effect cannot be explained away by unobserved confounding. In sum, we do not see evidence that unobserved confounders may undermine our estimates of treatment effectiveness, which is in line with many existing works using randomized control trials that have already confirmed the effectiveness of \emph{metformin} \citep[e.g.,][]{Knowler.2002}. 

\begin{table}[H]
\TABLE
{Results of robustness check for unobserved confounders.\label{tbl:unobs_conf}}
{\scriptsize
\begin{tabular}{l SSSSS}
\toprule
Strength & {All patients} & \multicolumn{4}{c}{{Subgroups}} \\
\cmidrule(l){3-6}
& & {A} & {B} & {C} & {D} \\
\midrule
$0.2 \times age$ & 0.066 & 0.066 & 0.142 & 0.048 & 0.058 \\
 & {[0.052, 0.088]} & {[0.045, 0.096]} & {[0.103, 0.188]} & {[0.026, 0.074]} & {[0.027, 0.097]} \\
$0.5 \times age$ & 0.053 & 0.052 & 0.127 & 0.037 & 0.046 \\
 & {[0.039, 0.074]} & {[0.029, 0.078]} & {[0.084, 0.177]} & {[0.015, 0.060]} & {[0.019, 0.075]} \\
$0.8 \times age$ & 0.043 & 0.040 & 0.117 & 0.029 & 0.039 \\
 & {[0.031, 0.058]} & {[0.013, 0.072]} & {[0.075, 0.132]} & {[0.009, 0.052]} & {[0.014, 0.071]} \\
\bottomrule
\multicolumn{6}{l}{Stated: mean estimation (95\% confidence intervals in parentheses)}
\end{tabular}}
{\SingleSpacedXI\footnotesize{\textit{Note.} Estimated treatment effects when assuming the existence of an unobserved confounder with the strength $0.2 \times age$, $0.5 \times age$, and $1.0 \times age$.}}
\end{table} 

\clearpage

\section{Machine Learning Explainability} 
\label{appendix:feature_importance}

To determine the most influential predictors, we calculated SHAP values. SHAP values are a unified method for measuring how much a predictor contributes to the overall model output and, therefore, rank the importance of the features. Moreover, they indicate whether higher (lower) values of a predictor are associated with an increased risk of diabetes onset.

\Cref{fig:feat_imp} shows the ten most important predictors for the gradient boosted decision tree when predicting diabetes onset. The most important predictors are age, body mass index (BMI), and the current glycated hemoglobin level (HbA1c). Age and BMI are well known risk factors for diabetes, and both were accordingly among the most important predictors for our gradient boosted decision tree \citep{abbasi2016systematic}. Also among the ten most important predictors are the red cell distribution width and thyroid stimulating hormone tests, which have both been found in the medical literature to be predictors for diabetes \citep[\eg,][]{chaker2016thyroid}. The height and weight of the patient are also among the ten most important predictors, yet with less importance. This is likely due to the fact that the predictive power of these variables as individual predictors is limited, and only their combination (as in the BMI) leads to an important predictor.  

We discussed our results with medical experts from diabetes care, who confirmed to us that these predictors are well-established risk factor for diabetes.    

\begin{figure}[H]
\FIGURE
{\includegraphics[width=.9\textwidth]{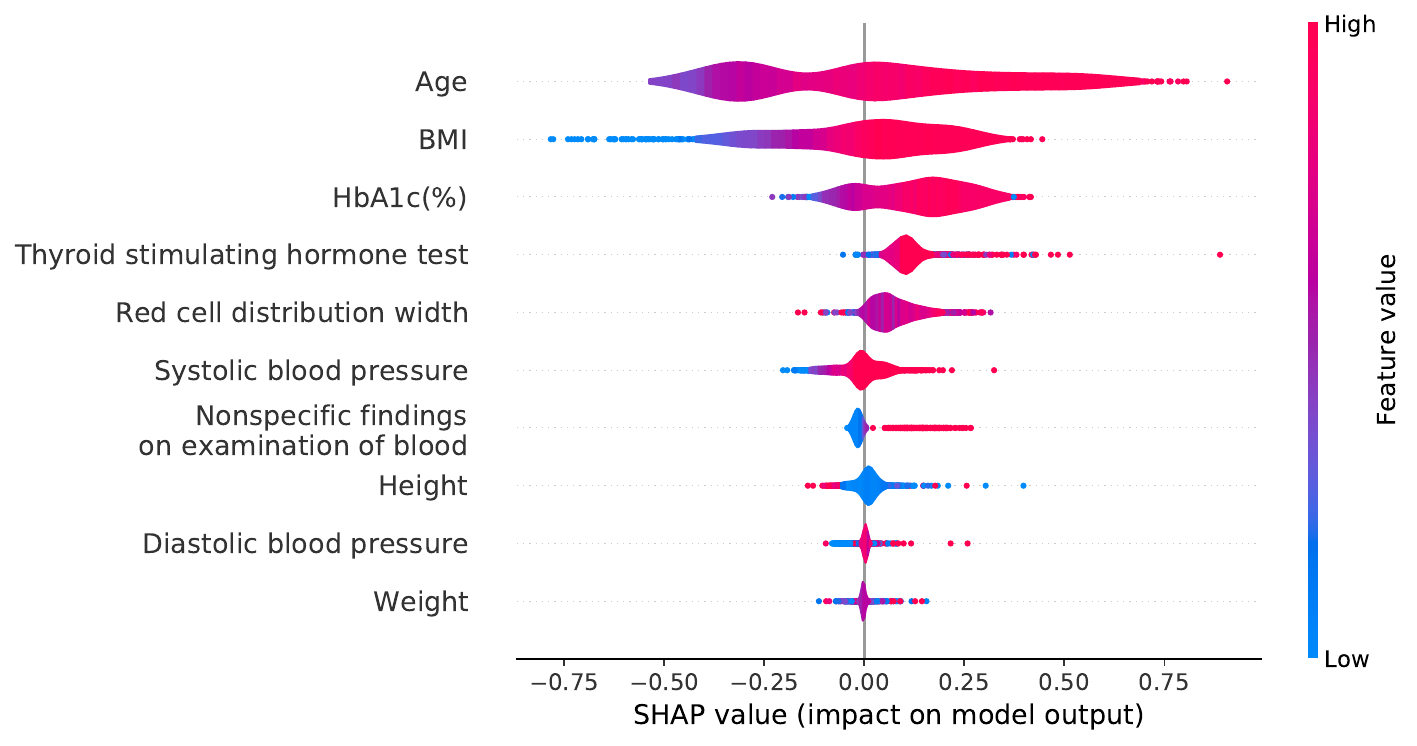}}
{SHAP values for the ten most important predictors for the prediction model. The dots represent observation for which the x-axis denotes the SHAP value. Observations with a positive (negative) SHAP value denote a higher (lower) risk for a diabetes onset. The color of each dot indicates the corresponding value of the predictor for a given observation. \label{fig:feat_imp}}
{}
\end{figure}

\clearpage

\section{Calibration of Machine Learning Model}
\label{appendix:model_calibration}

We use two techniques to improve the calibration of our machine learning model. Here, the aim is that the distribution of the predicted probability is similar to the distribution of the observed probability in the training data. First, we use the synthetic minority over-sampling technique (SMOTE) \citep{chawla2002smote} to oversample observations from patients who develop diabetes. Second, we use Platt scaling to calibrate the predicted class probabilities \citep{niculescu2005predicting}. \Cref{fig:calibration} shows the calibration plot for our gradient boosted decision trees after resampling. The calibration plot shows a good calibration. 

\begin{figure}[H]
\FIGURE
{\includegraphics[width=.45\textwidth]{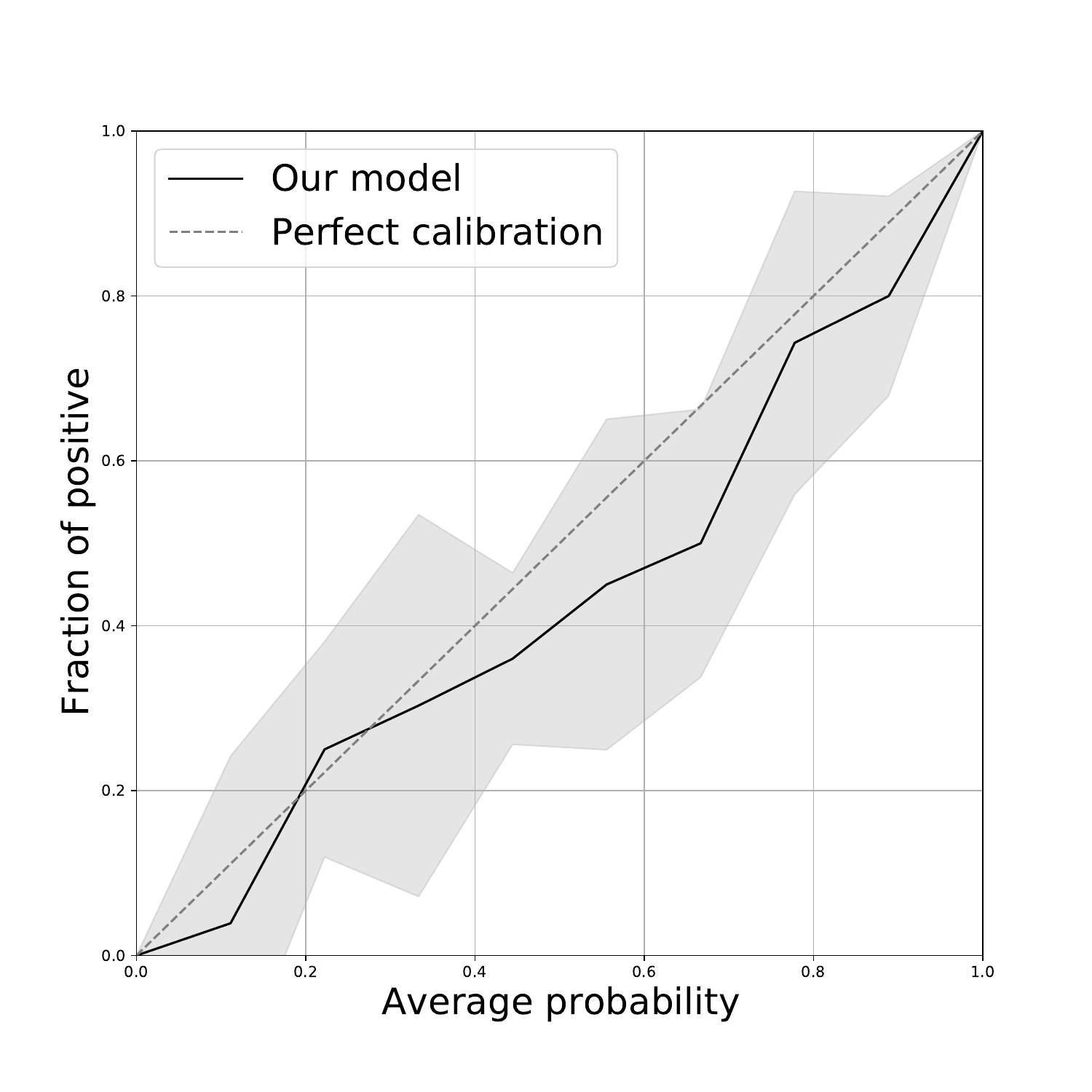}}
{Calibration of our machine learning model. The gray area shows the standard deviation due to resampling. \label{fig:calibration}}
{}
\end{figure}

\clearpage

\section{Hyperparameter Tuning}
\label{appendix:tuning_parameter}

We determined the hyperparameters using Bayesian optimization with 100 iterations of optimization. \Cref{tbl:tuning_parameter} lists the tuning ranges for each parameter. In the Bayesian optimization, the initial parameters are sampled from a uniform distribution. We implemented and evaluated all decision models in Python. The lasso, ridge regression, and random forest machine learning models are implemented using scikit-learn. The deep neural network is implemented using Pytorch. We train the deep neural network using the Adam optimizer. We use ReLU activation in the hidden layers and sigmoid in the output layer. The objective of the optimizer is to minimize the binary cross-entropy. The gradient boosted decision trees are implemented using LightGBM. For the causal forest, we keep all parameters at their default values. That is, we set the number of estimators to 10, the number of features to consider when looking for the best split to 10, the maximum depth of the trees to 5, the minimum number of samples required to be split at a leaf node to 100, and the minimum number of samples required of the experiment group to be split at a leaf node to 10.


\begin{table}[H]
\TABLE
{Hyperparameter tuning.\label{tbl:tuning_parameter}}
{\OneAndAHalfSpacedXI
	\centering
	\hspace{0.cm}
	\footnotesize
	\begin{tabular}{lll}
		\toprule
		\textbf{Model} & \textbf{Tuning parameters} & \textbf{Tuning range} \\ 
		\midrule
		Lasso & Regularization strength $\alpha$ & $(0.001, 10000)$ \\[0.5em]
		Ridge regression & Regularization strength & $(0.001, 10000)$ \\[0.5em]
		Random forest & Number of trees & $(20, 200)$ \\ 
		& Min samples for split & $(2, 150)$ \\
		& Number of features to consider for split & $\{ \textrm{sqrt, log2, all} \}$ \\
		& Criterion for split & $\{ \textrm{gini, entropy} \}$ \\
		& Class weight & $\{ \textrm{none, balanced} \}$ \\[0.5em]
		Deep neural network & Number of neurons & $(5, 128)$ \\ 
		& Number of hidden layers & $(1, 5)$ \\
		& Regularization strength & $(0.0, 5.0)$ \\
		& Learning rate & $(0.0001, 0.1)$ \\
		& Batch size & $(10, 1024)$ \\[0.5em]
		Gradient boosted decision trees & Number of trees & $(20, 200)$ \\ 
		& Number of leaves & $(20, 150)$ \\
		& Learning rate & $(0.01, 0.5)$ \\
		& Number of samples for constructing bins & $(20000, 300000)$ \\
		& Minimum number of data needed in a child & $(20, 500)$ \\
		& L1 regularization term on weights & $(0, 1)$ \\
		& L2 regularization term on weights & $(0, 1)$ \\
		& Class weight & $\{ \textrm{none, balanced} \}$ \\
		\bottomrule
	\end{tabular}}
{}
\end{table} 

\clearpage

\section{Applicability for Settings with Known Treatment Effects}
\label{appendix:personalized_coaching}

Here, we aim to demonstrate the applicability of our model in settings where the treatment effect of preventive care is known. In such a case, there is no need for healthcare organizations to perform counterfactual inference for treatment effect estimation, as the treatment effect can the be directly entered in the decision model.  

To show the applicability of our decision model, we build upon a different setting. Recall that we examined \emph{metformin} in the main paper. The reason was that it represents the quasi-standard for preventive care aimed at individuals at risk of developing diabetes in current medical practice \citep{Long.2016}. Several studies also suggest a risk reduction if patients are enrolled in personalized coaching toward a healthier lifestyle; however, such personalized coaching is currently not covered by health insurers in many countries (\eg, Germany, Switzerland, and Medicare in the United States). Nevertheless, it may be interesting to study the cost-effectiveness of lifestyle coaching for diabetes prevention.  

In the case of personalized coaching, the goal of preventive treatment is to achieve lifestyle changes of the patients. Lifestyle changes have not been prescribed to the customers of our partnering health insurer. Thus, the treatment effect cannot be estimated through stage~1 of our decision model. Instead, we rely on prior literature that has estimated the treatment effect of personalized coaching through randomized controlled trials \citep{Knowler.2002}. 

We assume a treatment effect due to personalized coaching of $\gamma_{i,l^*}=0.58$ \citep{Knowler.2002} that is exponentially decreasing over time, \ie,
\begin{equation}
    \gamma_{i,l} = \gamma_{i,l^*+1} \, \exp(l^*-l), \quad \text{for } l \geq l^*.
\end{equation}
We set the annual cost of personalized coaching to $C_{\mathrm{prevent}} = 1,600$ \citep{azelton2021digital}.

\Cref{tbl:res_vs_current_app} shows the empirical results. Here, we again compare the data-driven allocations from our decision model against that from current practice. Our findings are line with the results from the main paper: our decision model outperforms current practice across all metrics. 

To sum up, there are different benefits of whether the treatment effect is (a)~known or (b)~estimated through stage~1 of our decision model. A benefit of (a) is that healthcare organizations can leverage known treatment effects from randomized controlled trials, which represent the gold standard for measuring treatment effects. In particular, there is no bias due to unobserved confounders. A benefit of (b) is that randomized controlled trials are not always available or otherwise costly, because of which our estimation may be preferred in practice. Moreover, using counterfactual inference in our decision model offers a mathematical approach to directly learn heterogeneous treatment effects and thus account for the differential effectiveness of preventive care across patients.  

\begin{table}[H]
\TABLE
{Comparison of current practice vs. our data-driven decision model for the case of personalized coaching.\label{tbl:res_vs_current_app}}
{\footnotesize
\begin{tabular}{l SSS SSS}
\toprule
& \multicolumn{2}{c}{$k$=1,000} & \multicolumn{2}{c}{$k$=5,000} & \multicolumn{2}{c}{$k$=10,000} \\
\cmidrule(l){2-3} \cmidrule(l){4-5} \cmidrule(l){6-7} 
& {Prevented} & {Cost} & {Prevented} & {Cost} & {Prevented} & {Cost} \\
& {diseases} & {savings} & {diseases} & {savings} & {diseases} & {savings} \\
\midrule
\textbf{Na{\"i}ve baseline}  &       728.378  &              16.457  &       1423.607  &               26.651  &       2736.326  &               45.807  \\
          &         (5.606)  &               (0.141)  &         (10.761)  &                (0.249)  &         (16.553)  &                (0.323)  \\
\textbf{Current practice}  &      1318.303  &              25.786  &       2083.693  &               37.088  &       3184.516  &              51.263   \\
          &         (9.893)  &               (0.223)  &         (12.719)  &                (0.287)  &         (14.837)  &                (0.310)  \\
\midrule
\textbf{Our decision model}  &      1555.792  &              29.544  &       2143.867  &               38.036  &       3349.127  &               53.767  \\
          &        (10.960)  &               (0.223)  &         (14.553)  &                (0.295)  &         (15.103)  &                (0.276)  \\
\bottomrule
\multicolumn{7}{l}{Stated: mean performance (standard deviation in parentheses)}
\end{tabular}}
{\SingleSpacedXI\footnotesize{\textit{Note.} Performance metrics for allocating preventive care given a varying budget for enrolling $k$ patients per year into preventive treatments. Cost savings over no preventive care allocation are reported in USD millions.}}
\end{table} 

\end{APPENDICES}

\end{document}